\documentclass[10pt,twocolumn,letterpaper]{article}

\usepackage[pagenumbers]{cvpr} 

\usepackage{graphicx}
\usepackage{subcaption}
\usepackage{amsmath}
\usepackage{amssymb}
\usepackage{mathtools}
\usepackage{amsthm}
\usepackage{array}
\usepackage{multirow}
\definecolor{darkgreen}{rgb}{0.0, 0.5, 0.0}
\usepackage{pgfplots}
\usepackage{pgfplotstable}
\usepackage{makecell}

\usepackage{setspace}

\definecolor{cvprblue}{rgb}{0.21,0.49,0.74}
\usepackage[pagebackref,breaklinks,colorlinks,allcolors=cvprblue]{hyperref}

\def\paperID{12160} 
\def\confName{CVPR}
\def\confYear{2026}

\title{Not All Attention Heads Are What You Need: Refining CLIP's Image Representation with Attention Ablation}

\author{Feng Lin\\
Intellifusion Inc.\\
\and
Marco Chen\\
Intellifusion Inc.\\
\and
Haokui Zhang \\
Northwest Polytechnical University \\
\and
Xiaotian Yu \\
Intellifusion Inc. \\
\and
Guangming Lu \\
Harbin Institute of Technology \\
\and
Rong Xiao \\
Intellifusion Inc. \\
}

\begin{document}
\maketitle
\begin{abstract}
This paper investigates the role of attention heads in CLIP's image encoder. Building on interpretability studies, we conduct an exhaustive analysis and find that certain heads, distributed across layers, are detrimental to the resulting representations. To mitigate their impact, we propose a simple yet effective Attention Ablation Technique (AAT) that suppresses selected heads by directly manipulating their attention weights. By incorporating two complementary strategies tailored to different application scenarios,  AAT enables the systematic identification and ablation of harmful heads with minimal overhead. Experiments show that AAT consistently improves downstream performance across diverse domains, boosting recall by up to 11.1\% on cross-modal retrieval benchmarks. These results highlight that AAT can effectively refine large-scale VLMs with virtually no extra inference cost, while yielding semantically meaningful patterns that align with existing interpretability findings.\end{abstract}    
\section{Introduction}\label{sec:intro}
As a pioneering large-scale vision-language model (VLM), CLIP~\cite{radford2021learning} has garnered widespread attention for its simple yet effective design and impressive capability across a wide range of downstream tasks~\cite{jia2021scaling, zhou2022conditional, luddecke2022image, cho2021unifying}. Early CLIP research focuses on advances in data~\cite{li2021align}, supervision~\cite{zhai2023sigmoid}, and architecture~\cite{jia2021scaling}, recent studies have shifted toward analyzing its learned representations~\cite{gandelsmaninterpreting, lan2025clearclip, abbasi2025deciphering}. These efforts aim to uncover the intrinsic characteristics of CLIP's representations, providing insights for further enhancement.

\begin{figure}[t]
\centering
\includegraphics[width=3.2in]{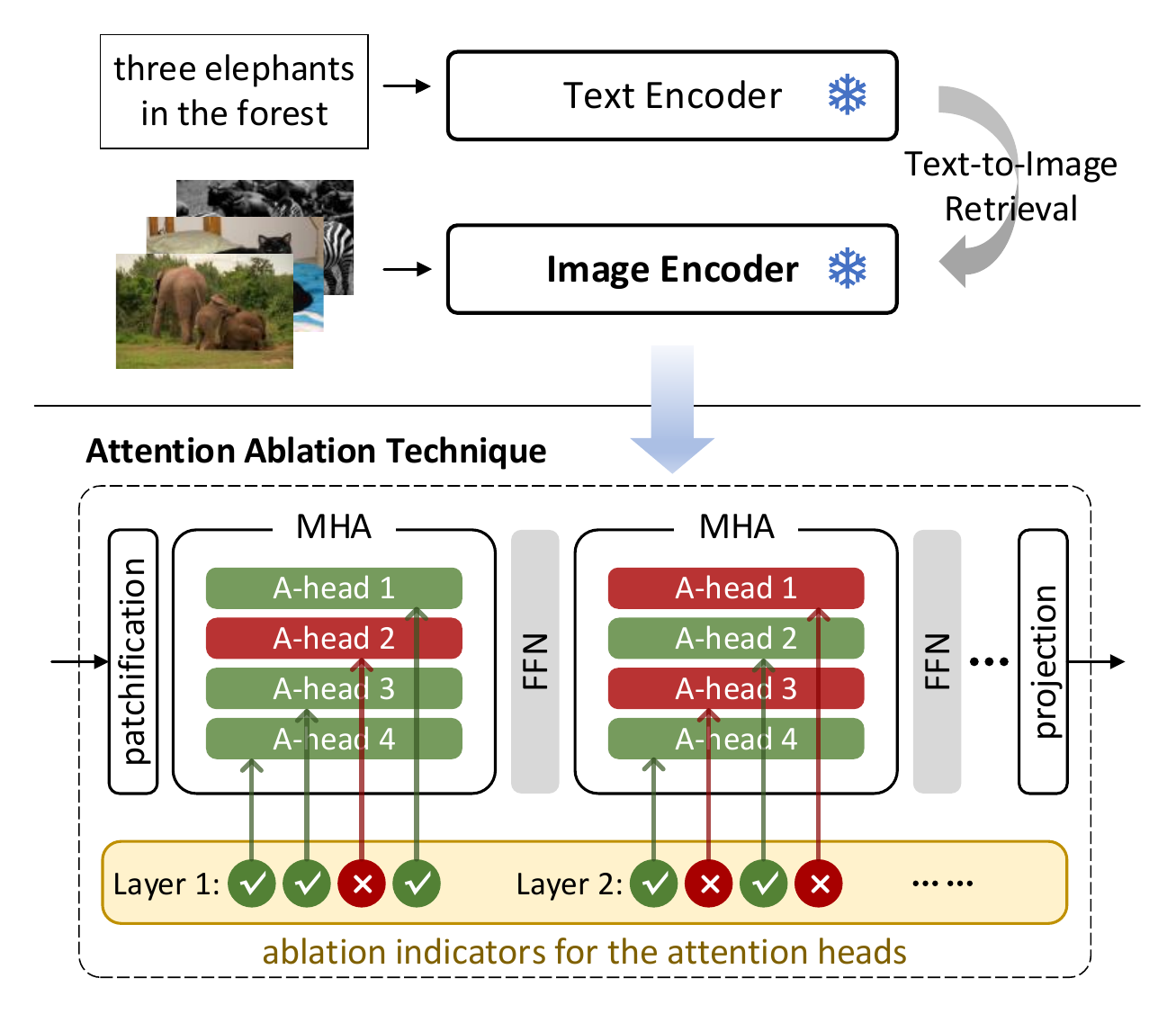}
\vspace{-7.5pt}
\caption{An illustration of AAT-improved CLIP for text-to-image retrieval. ``A-head $i$'' denotes the $i$-th head in MHA. With model weights frozen, AAT ablates the selected image encoder's heads. Cross marks denote head ablation while check marks for retention.}
\vspace{-10pt}
\label{fig:overview}
\end{figure}

A recent interpretability study~\cite{gandelsmaninterpreting} investigates CLIP's image representations by decomposing them across attention heads, revealing strong correlations between specific heads and particular linguistic concepts. Through manually removing those heads associated with spurious cues in the last four transformer layers, it improves CLIP's image representations for a targeted zero-shot classification task. This finding suggests that individual head makes distinct contribution to output representations, acting as property-specific ``filters'' over visual concepts. Building on this insight, we systematically extend the analysis to all attention heads in CLIP's image encoder across diverse downstream tasks.

Trained on vast amounts of internet-sourced image–text pairs, CLIP models probably develop attention heads that encode task-irrelevant signals---such as domain biases or spurious cues---that may hinder downstream performance. Some heads may also overfit to noise inherent in uncurated training data. Such detrimental heads can appear throughout the transformer layers of the image encoder. We hypothesize that ablating them can refine the output and thereby yield consistent gains across diverse downstream tasks.

To verify our hypothesis, we conduct a preliminary experiment to assess the impact of individual attention heads. We find that ablating even a single detrimental head can enhance CLIP's downstream performance, with further gains from ablating simple combinations of negatively impactful heads. Inspired by this, we develop an efficient Attention Ablation Technique (AAT) that refines the representations by evicting groups of detrimental heads. It formulates the problem as a combinatorial optimization task, solved using either Genetic Algorithm (GA) or Back-Propagation (BP).

The key difference between GA-based AAT (AAT-GA) and BP-based AAT (AAT-BP) lies in how they identify detrimental attention heads. AAT-GA uses a fitness function that measures the distance between positive image–text pairs and hard negative pairs, whereas AAT-BP introduces a lightweight training scheme with a learnable gating parameter for each head to assess its relevance. Both strategies yield substantial gains on multi-domain retrieval and classification tasks, with their respective strengths making them suitable for different deployment scenarios.

AAT leaves CLIP's parameters unchanged and instead operates by manipulating the attention weights of selected heads, as illustrated in Figure~\ref{fig:overview}. It suppresses the weights on image tokens while correspondingly amplifying the class token weight, ensuring that ablated heads contribute minimally to the output representation. As an agile and versatile technique, AAT is particularly well suited to resource-constrained scenarios, like limited compute or scarce high-quality data, while preserving strong generalization across diverse domains. The main contributions as follows:
\begin{itemize}
\item We extensively explore the impacts of attention heads in CLIP's image encoder and show that ablating detrimental heads improves representation quality. To our knowledge, this is the first systematic study of refining CLIP's outputs by probing the impact of individual attention heads.
\item We introduce AAT, a simple yet effective method for automatically identifying and ablating detrimental attention heads without altering original model weights, offering two variants tailored to different application scenarios.
\item AAT delivers consistent and substantial gains across diverse downstream tasks, improving recall by up to 11.1\% on retrieval benchmarks with negligible extra overhead.
\item Out result analysis provides concrete evidence that attention heads are semantically meaningful, reinforcing and extending existing interpretability findings of VLMs.
\end{itemize}
\section{Related Work}
\paragraph{The Interpretability of CLIP's Representation}
Recent studies investigate CLIP's representations from various perspectives. \cite{materzynska2022disentangling} analyzes entanglement between words and images. \cite{li2023clip} presents a semantic shift toward background regions. \cite{shi2023towards} finds a strong connection between the modality gap and the loss local minima. \cite{zhao2024gradient} improves image-text matching explainability via gradients and heatmaps. \cite{crabbeinterpreting} finds that CLIP exhibits distinctive outlier features tied to ImageNet shift robustness. Most closely related to AAT, \cite{bhalla2024interpreting} boosts interpretability by converting dense embeddings into sparse, concept-based ones, and \cite{gandelsmaninterpreting} reveals spatial localization and property-specific roles of attention heads.

\paragraph{Attention Weight Manipulation}
Attention mechanism is central to Transformers~\cite{brauwers2021general}, and manipulating attention weights in trained models has shown notable benefits. Prior heuristic method calibrates its weights toward hidden attention sinks in LLMs~\cite{yu2024unveiling}; reformulates attention for segmentation~\cite{lan2025clearclip}; reweights semantic tokens to refine text embeddings~\cite{kim2024semantic}; and amplifies image-token weights to mitigate hallucinations in VLMs~\cite{liu2025paying}. AAT is conceptually related to them but introduces a novel, empirically grounded manipulation strategy based on attention-head behaviors.

\paragraph{Attention Head Pruning in LMs} 
The idea behind AAT parallels early works on language models~\cite{michel2019sixteen, voita2019analyzing, behnke2020losing}, which reveal redundancy among attention heads. Those methods typically prune heads to slim models with modest accuracy loss. Though head pruning is well established for LMs, its study to much compact VLMs remains underexplored.

\paragraph{Parameter-Efficient Finetuning}
AAT-BP is superficially similar with PEFT~\cite{hulora, lester2021power, houlsby2019parameter, sung2022lst, wu2024reft}---requiring only hundreds of learnable gates---but, unlike PEFT's largely black-box updates, each parameter in AAT-BP is explicitly interpretable. It is also far more parameter-efficient, requiring orders of magnitude fewer parameters than PEFT~\cite{ding2023parameter}.
\begin{table}[t]
\centering
\small
\begin{tabular}{p{30pt}<{\centering}|p{19pt}<{\centering}p{19pt}<{\centering}p{19pt}<{\centering}p{19pt}<{\centering}p{19pt}<{\centering}|p{25pt}<{\centering}}
\hline
indices & 11-6 & 10-7 & 10-9 & 9-9 & 11-0 & vanilla \\
\hline
mean-R & 81.1 & 80.8 & 80.8 & 80.7 & 80.7 & 79.9 \\
\hline
\end{tabular}
\vspace{-5pt}
\caption{Top five single-head ablation configurations for text-to-image retrieval on COCO-CN $val$ set~\cite{li2019coco}. ``\emph{m-n}'' demotes the $n$-th head in the $m$-th layer. Evaluations are based on CLIP-ViT-B~\cite{yang2022chinese}, mean-R (the mean of R@1, R@5, and R@10) as metric.}\label{tab:pre1}
\end{table}

\begin{table}[t]
\centering
\small
\begin{tabular}{p{100pt}p{30pt}<{\centering}p{30pt}<{\centering}p{30pt}<{\centering}}
\hline
Models & $val$ & $test$ & $all$ \\
\hline
vanilla CLIP~\cite{yang2022chinese} & 79.9 & 81.1 & 41.7 \\
naive joint head ablation & 82.5 & 82.6 & 43.2 \\
\hline
\end{tabular}
\vspace{-5pt}
\caption{Comparison of mean-R for text-to-image retrieval on the COCO-CN $val$, $test$ and $all$ set.}
\vspace{-5pt}
\label{tab:pre2}
\end{table}

\section{The Impact of Individual Attention Heads}\label{sec:gridsearch}
As a preliminary step, we conduct a simple head-wise ablation sweep of CLIP's image encoder. Using AAT (described in Section~\ref{sec:aat}), we ablate one attention head at a time and evaluate cross-modal retrieval performance. For ViT-B with 144 heads across 12 layers, this yields 144 results, several of which outperform the original baseline. The top 5 highest-performing configurations are ranked in Table~\ref{tab:pre1}, with full results in Appendix~\ref{app:greedsearch}. Remarkably, ablating even a single detrimental head can raise mean recall by up to 1.2\%.

Building on this observation, we ablate the direct union of all the individually identified detrimental heads---those whose single-head ablation improves recall. As shown in Table~\ref{tab:pre2}, this naive combination consistently boosts retrieval performance across multiple test subsets. Notably, although these heads are selected independently, their joint ablation yields further gains. This motivates us to develop more principled strategies for optimal head selection.

\section{Attention Ablation Technique}
We introduce AAT in this section. Section~\ref{sec:aat} explains the method for ablating individual attention heads. Sections~\ref{sec:aat-ga} and~\ref{sec:aat-bp} present two alternative strategies---GA and BP---for identifying a globally optimal set of detrimental heads, with each strategy tailored to different application scenarios.

\subsection{Ablating a Single Attention Head}\label{sec:aat}
A transformer layer consists of a multi-head attention module (MHA) and a feed-forward network (FFN)~\cite{dosovitskiy2020image}. For the CLIP's image encoder, the input token sequence of length $N$ is composed of a leading class token followed by $N-1$ image tokens.
Let $x^{l}_{h}$ represent the input to the $h$-th attention head of MHA in the $l$-th transformer layer. The output of each head, ${y}^{l}_{h}$, can be formulated as follows:
\begin{align}
\mathcal{A}^{l}_{h} &= Softmax(\frac{f_{q}(LN(x^{l}_{h})) \cdot f_{k}(LN(x^{l}_{h})^T)}{\sqrt{d_k}}), \\
{y}^{l}_{h} &= \mathcal{A}^{l}_{h} \cdot f_{v}(LN(x^{l}_{h}))
\end{align}
where $f_{q}$, $f_{k}$, and $f_{v}$ denote projection layers for queries, keys, and values, respectively. $LN$ refers to LayerNorm, $d_{k}$ is the embedding dimension of each head, and $\mathcal{A}^{l}_{h}$ denotes the attention weight matrix with dimensions $N \times N$.

\begin{figure}[t]
\centering
\includegraphics[width=2.7in]{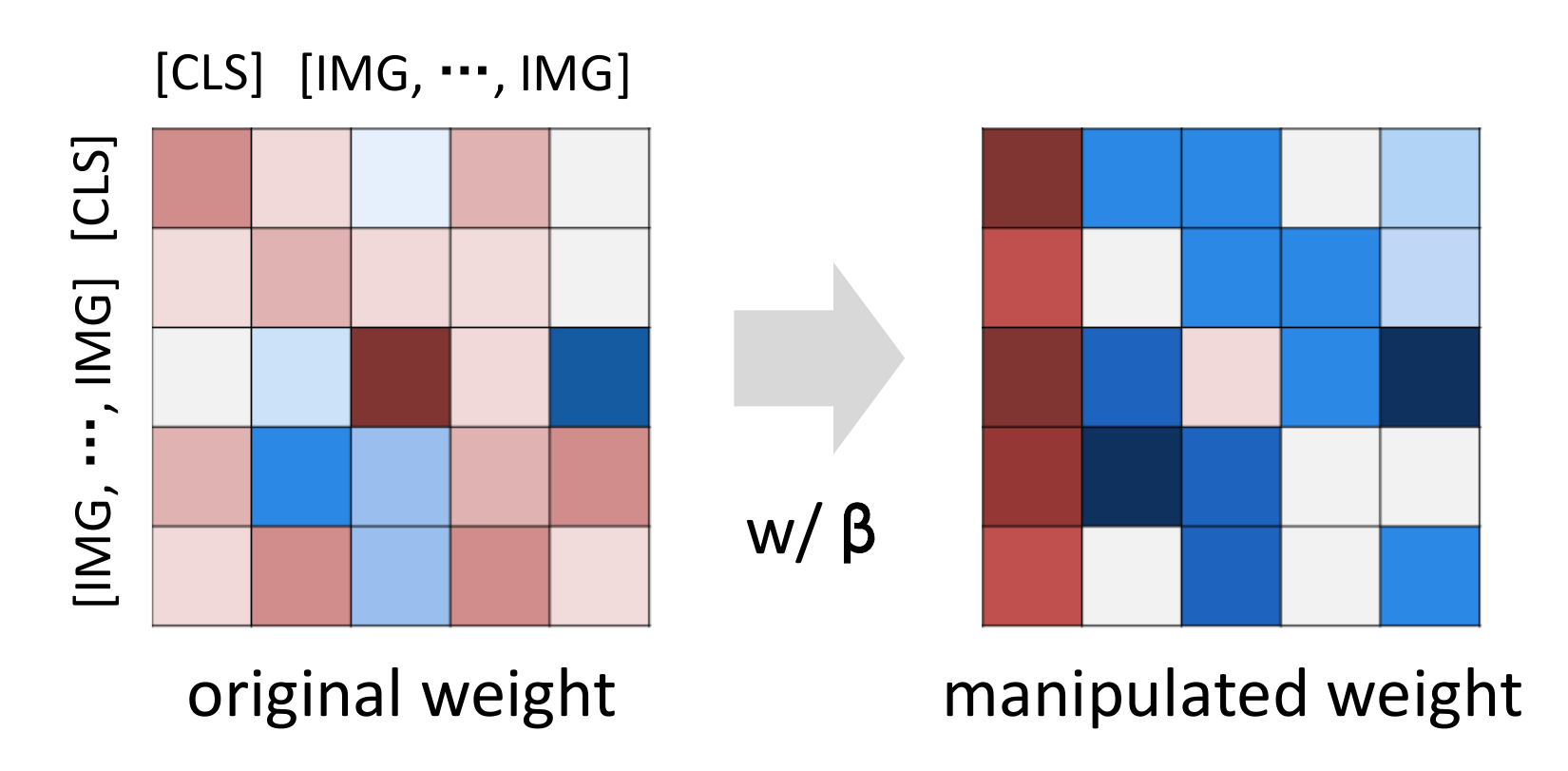}
\vspace{-5pt}
\caption{The left $5 \times 5$ matrix shows the original attention weight, while the right depicts it after manipulation. CLS and IMG denote the class token and an image token, respectively, with a total length of 5. Darker blue indicates lower attention scores (approaching zero), while darker red indicates higher scores (approaching one).}
\vspace{-5pt}
\label{fig:aat}
\end{figure}

As shown in the formulations above, the attention weight matrix $\mathcal{A}$ is produced by a softmax function, ensuring that each row sums to 1. The output of each attention head, ${y}$, is then obtained by multiplying $\mathcal{A}$ with the projected values. To ablate a specific head, we manipulate $\mathcal{A}$ to suppress the contribution of image tokens to the output representation, as illustrated in Figure~\ref{fig:aat}. Specifically, this involves two steps:
\begin{enumerate}
\item Reducing the importance of image tokens: Modify $\mathcal{A}$ as $\mathcal{A}[:,j] \leftarrow \mathcal{A}[:,j] \times \beta$, where $j \in \{1, 2, ..., N-1\}$.
\item Renormalizing each row of $\mathcal{A}$: Scale $\mathcal{A}$ as $\mathcal{A}[i,j] \leftarrow \mathcal{A}[i,j] / \mathcal{A}_i$, where $i,j \in \{0, 1, ..., N-1\}$ and $\mathcal{A}_i = \sum_j \mathcal{A}[i,j]$, to preserve the row-wise normalization.
\end{enumerate}
A hyper-parameter $\beta$ controls the degree of suppression applied to image-token weights; unless otherwise stated, we set $\beta = 0.1$. After these two steps, the modified attention weight replaces the original, while all other MHA operations remain unchanged. This allows refinement of the output representation without altering the model parameters.

\begin{table*}[ht]
\centering
\fontsize{7}{9}\selectfont
\begin{tabular}{p{10pt}<{\centering}|p{35pt}<{\centering}|p{12pt}<{\centering}p{12pt}<{\centering}p{12pt}<{\centering}p{12pt}<{\centering}|p{12pt}<{\centering}p{12pt}<{\centering}p{12pt}<{\centering}p{12pt}<{\centering}|p{12pt}<{\centering}p{12pt}<{\centering}p{12pt}<{\centering}p{12pt}<{\centering}|p{12pt}<{\centering}p{12pt}<{\centering}p{12pt}<{\centering}p{12pt}<{\centering}}
\hline
\multicolumn{2}{c|}{tasks} & \multicolumn{8}{c|}{text-to-image retrieval} & \multicolumn{8}{c}{image-to-text retrieval} \\
\hline
\multicolumn{2}{c|}{splits} & \multicolumn{4}{c|}{the \emph{test} set} & \multicolumn{4}{c|}{the \emph{all} set} & \multicolumn{4}{c|}{the \emph{test} set} & \multicolumn{4}{c}{the \emph{all} set} \\
\hline
\multicolumn{1}{c|}{size} & \multicolumn{1}{c|}{method} & R@1 & R@5 & R@10 & mR & R@1 & R@5 & R@10 & mR & R@1 & R@5 & R@10 & mR & R@1 & R@5 & R@10 & mR \\
\hline
\multirow{3}{*}{B} & vanilla & 38.8 & 64.2 & 73.8 & 58.9 & 13.0 & 26.5 & 33.8 & 24.4 & 57.8 & 81.7 & 88.6 & 76.0 & 24.2 & 43.1 & 52.1 & 39.8 \\
 & AAT-GA & \textbf{41.3} & \textbf{66.7} & \textbf{76.6} & \textbf{61.5} & \textbf{14.0} & \textbf{28.3} & \textbf{35.9} & \textbf{26.1} & \textbf{60.3} & \textbf{82.7} & 89.3 & \textbf{77.4} & \textbf{26.0} & \textbf{45.4} & \textbf{54.4} & \textbf{41.9} \\
 & AAT-BP & 40.5 & 66.2 & 75.9 & 60.9 & 13.9 & 28.1 & 35.6 & 25.9 & 58.9 & 82.5 & \textbf{89.5} & 77.0 & 24.8 & 44.2 & 53.1 & 40.7 \\
\hline
\multirow{3}{*}{L} & vanilla & 43.9 & 68.8 & 78.3 & 63.7 & 16.1 & 31.3 & 39.1 & 28.8 & 59.5 & 82.3 & 89.2 & 77.0 & 25.5 & 44.9 & 53.9 & 41.4 \\
 & AAT-GA & \textbf{46.3} & 71.3 & 80.8 & 66.1 & \textbf{17.4} & \textbf{33.3} & \textbf{41.4} & \textbf{30.7} & \textbf{62.8} & \textbf{84.2} & \textbf{90.2} & \textbf{79.1} & \textbf{27.6} & \textbf{47.9} & \textbf{56.8} & \textbf{44.1} \\
 & AAT-BP & 46.0 & \textbf{71.8} & \textbf{80.9} & \textbf{66.2} & 17.2 & 33.2 & 41.3 & 30.6 & 60.6 & 83.3 & 89.7 & 77.8 & 25.6 & 45.6 & 54.7 & 42.0 \\
\hline
\multirow{3}{*}{H} & vanilla & 47.4 & 72.1 & 80.9 & 66.8 & 18.5 & 34.8 & 42.9 & 32.1 & 60.6 & 84.6 & 90.9 & 78.7 & 26.0 & 46.1 & 55.4 & 42.5 \\
 & AAT-GA & \textbf{49.3} & \textbf{74.1} & \textbf{82.7} & \textbf{68.7} & \textbf{19.5} & \textbf{36.5} & \textbf{44.7} & \textbf{33.6} & \textbf{63.8} & 85.8 & 91.5 & 80.4 & \textbf{28.2} & 48.7 & \textbf{58.1} & \textbf{45.0} \\
 & AAT-BP & 48.6 & 73.5 & 82.2 & 68.1 & 19.1 & 35.9 & 44.1 & 33.0 & 63.7 & \textbf{85.9} & \textbf{91.5} & \textbf{80.4} & 28.1 & \textbf{48.8} & 58.0 & 45.0 \\
\hline
\end{tabular}
\vspace{-2pt}
\caption{Comparison on MS COCO for cross-modal retrieval. R@k denotes recall rate at rank k, and mR (mean-R) denotes the average of R@1, R@5, and R@10. Model sizes are denoted as B (ViT-B), L (ViT-L), and H (ViT-H). Best results are highlighted in bold.}\label{tab:coco-t2i-i2t}
\end{table*}

\subsection{AAT with Genetic Algorithm}\label{sec:aat-ga}
As discussed in Section~\ref{sec:gridsearch}, ablating detrimental heads in CLIP's image encoder can refine its output representations. Our goal is to identify the optimal subset of such heads. We formulate this as a combinatorial optimization problem and solve it using a small validation set $\mathcal{D}$ containing a limited number of image–text pairs. To efficiently explore the large search space, we employ a Genetic Algorithm (GA)~\cite{holland1992adaptation}, which evolves candidate solutions over generations to discover high-quality configurations.

Specifically, we represent all attention heads in the image encoder as a binary vector, where each element corresponds to a head (1 for ablated, 0 for retained). Using the validation set $\mathcal{D}$, the GA optimizes this vector configuration. To align with CLIP's training objective which brings matching image–text pairs closer while pushing mismatched pairs apart, we design a fitness function $\mathcal{F}$ that reflects this property. $\mathcal{F}$ serves as the optimization target and is defined as follows:
\begin{align}\label{fml:fit}
\mathcal{F} &= \frac{1}{N}\sum_{i=1}^{N}(S_{pos}^{i}-max_{j \in \mathcal{H}^{i}}(S_{neg}^{i,j}))
\end{align}
where $N$ is the number of sample pairs in $\mathcal{D}$. $S_{pos}^{i}$ denotes the cosine similarity between the $i$-th text and its ground-truth image. $S_{neg}^{i,j}$ denotes the cosine similarity between the $i$-th text and the $j$-th image in $\mathcal{H}^{i}$ (an updatable set of hard negatives for the $i$-th text), which is detailed below.

We first evaluate vanilla CLIP on $\mathcal{D}$ for text-to-image retrieval. For each text query $i$, we collect the top $k_1$ non-matching images (false positives) as hard negatives to initialize $\mathcal{H}^{i}$. To avoid overfitting, we augment $\mathcal{H}^{i}$ with $k_2$ randomly sampled non-matching images from $\mathcal{D}$ and refresh these $k_2$ samples at the end of each GA generation. This dynamic $\mathcal{H}$ is crucial to AAT-GA: it encourages larger margins between true matches and evolving hard negatives while continuously incorporating newly challenging cases.

In Eq.~\ref{fml:fit}, the first term of $\mathcal{F}$ rewards similarity between true image–text pairs, while the second penalizes similarity to the hardest negatives in $\mathcal{H}$. Combined with standard crossover and mutation, this objective steers the GA toward an optimal (or near-optimal) head selection for ablation.
\paragraph{Application Scenarios}
AAT-GA operates entirely at inference, making it well suited to resource-constrained deployments. It can run on the low-power, inference-oriented edge devices with limited-precision compute capabilities, such as INT8/INT16 chips (\emph{e.g.}, Google Coral Edge TPU, Ethos-N78, Mythic M1108 AMP). It is also data-efficient, requiring only a small validation set for optimization (Section~\ref{sec:dsize}), whereas supervised finetuning (SFT) tends to overfit in such settings (Appendix~\ref{app:sft}).

\subsection{AAT with Back-Propagation}\label{sec:aat-bp}
Since GA may require more inference trials when poorly initialized, in addition, not all resource-constrained settings lack floating-point compute, thus we introduce a gradient-based alternative: Back-Propagation (AAT-BP). Rather than using the uniform suppression factor $\beta$ in AAT-GA, AAT-BP assigns each probed head a learnable parameter $\alpha_i$. We map $\alpha_i$ to a head-specific suppression factor $\beta_i \in [0, 1]$ that controls the ablation strength of head $i$, replacing $\beta$ with $\beta_i$. Formally, $\beta_i$ is defined as:
\begin{align}
\beta_{i} &= \text{Sigmoid}(\tau \cdot \alpha_{i})
\end{align}
where the temperature $\tau>1$ sharpens the sigmoid, encouraging $\beta_i$ to approach extreme values (0 or 1) during optimization. Empirically, we set $\tau=5.0$ and initialize each $\alpha_i$ to 1.0, yielding $\beta_i$ near 1 at the start. As in AAT-GA, we optimize ${\alpha_i}$ on the same validation set $\mathcal{D}$ using CLIP's contrastive supervision objective.

\paragraph{Application Scenarios}
While BP might get stuck to local optima, its rapid convergence makes it practical and efficient in real-world use. Its low compute and data demands further suit resource-limited settings. In spirit, AAT-BP resembles PEFT, yet it uses orders of magnitude fewer learnable parameters than typical PEFT~\cite{hulora, houlsby2019parameter}, a level of efficiency enabled by targeted manipulation of attention heads.
\begin{table*}[ht]
\centering
\fontsize{7}{9}\selectfont
\begin{tabular}{p{10pt}<{\centering}|p{35pt}<{\centering}|p{12pt}<{\centering}p{12pt}<{\centering}p{12pt}<{\centering}p{12pt}<{\centering}|p{12pt}<{\centering}p{12pt}<{\centering}p{12pt}<{\centering}p{12pt}<{\centering}|p{12pt}<{\centering}p{12pt}<{\centering}p{12pt}<{\centering}p{12pt}<{\centering}|p{12pt}<{\centering}p{12pt}<{\centering}p{12pt}<{\centering}p{12pt}<{\centering}}
\hline
\multicolumn{2}{c|}{tasks} & \multicolumn{8}{c|}{text-to-image retrieval} & \multicolumn{8}{c}{image-to-text retrieval} \\
\hline
\multicolumn{2}{c|}{splits} & \multicolumn{4}{c|}{the \emph{test} set} & \multicolumn{4}{c|}{the \emph{all} set} & \multicolumn{4}{c|}{the \emph{test} set} & \multicolumn{4}{c}{the \emph{all} set} \\
\hline
\multicolumn{1}{c|}{size} & \multicolumn{1}{c|}{method} & R@1 & R@5 & R@10 & mR & R@1 & R@5 & R@10 & mR & R@1 & R@5 & R@10 & mR & R@1 & R@5 & R@10 & mR \\
\hline
\multirow{3}{*}{B} & vanilla & 67.20 & 88.68 & 93.06 & 82.98 & 30.54 & 52.17 & 61.27 & 47.99 & 85.90 & \textbf{97.90} & \textbf{99.10} & 94.30 & 48.16 & 72.76 & 81.18 & 67.36 \\
 & AAT-GA & \textbf{70.26} & 90.06 & \textbf{94.42} & \textbf{84.91} & 32.64 & \textbf{54.90} & \textbf{64.04} & \textbf{50.53} & \textbf{87.20} & 97.10 & 98.90 & \textbf{94.40} & \textbf{50.47} & \textbf{74.09} & 81.89 &\textbf{68.82} \\
 & AAT-BP & 69.76 & \textbf{90.08} & 93.96 & 84.60 & \textbf{32.69} & 54.79 & 63.98 & 50.49 & 86.30 & 97.70 & 99.00 & 94.33 & 49.58 & 73.77 & \textbf{81.97} & 68.44 \\
\hline
\multirow{3}{*}{L} & vanilla & 73.52 & 91.84 & 95.36 & 86.91 & 36.26 & 58.98 & 67.91 & 54.39 & 87.00 & 98.20 & \textbf{99.50} & 94.90 & 48.90 & 74.39 & 83.09 & 68.79 \\
 & AAT-GA & 75.74 & 92.72 & 95.92 & 88.13 & 38.28 & 61.16 & 69.93 & 56.46 & \textbf{88.10} & \textbf{98.50} & 99.40 & \textbf{95.33} & \textbf{52.68} & \textbf{76.96} & \textbf{84.90} & \textbf{71.51} \\
 & AAT-BP & \textbf{76.20} & \textbf{93.12} & \textbf{95.96} & \textbf{88.43} & \textbf{38.47} & \textbf{61.75} & \textbf{70.62} & \textbf{56.95} & 87.50 & 98.10 & 99.40 & 95.00 & 50.89 & 75.71 & 83.70 & 70.10 \\
\hline
\multirow{3}{*}{H} & vanilla & 76.10 & 93.44 & 96.42 & 88.65 & 41.18 & 64.03 & 72.61 & 59.27 & 88.60 & 98.70 & 99.60 & 95.63 & 51.10 & 77.44 & 85.64 & 71.39 \\
 & AAT-GA & \textbf{77.84} & \textbf{94.50} & \textbf{96.94} & \textbf{89.76} & \textbf{43.10} & \textbf{66.20} & \textbf{74.51} & \textbf{61.27} & 90.40 & 98.90 & 99.60 & 96.30 & 54.98 & 80.06 & 87.51 & 74.18 \\
 & AAT-BP & 77.78 & 94.48 & 96.92 & 89.73 & 42.62 & 65.78 & 74.32 & 60.91 & \textbf{90.70} & \textbf{98.90} & \textbf{99.60} & \textbf{96.40} & \textbf{56.16} & \textbf{80.79} & \textbf{88.05} & \textbf{75.00} \\
\hline
\end{tabular}
\vspace{-2pt}
\caption{Comparison on Flickr30k for cross-modal retrieval. The symbols in the table have the same meaning as those in Table~\ref{tab:coco-t2i-i2t}.}\label{tab:flickr30k-t2i-i2t}
\end{table*}

\begin{table*}[ht]
\centering
\fontsize{8}{10}\selectfont
\begin{tabular}{p{45pt}<{\centering}|p{16pt}<{\centering}p{16pt}<{\centering}p{16pt}<{\centering}p{16pt}<{\centering}p{16pt}<{\centering}p{16pt}<{\centering}p{16pt}<{\centering}p{16pt}<{\centering}p{16pt}<{\centering}|p{18pt}<{\centering}|p{16pt}<{\centering}p{16pt}<{\centering}p{16pt}<{\centering}|p{18pt}<{\centering}}
\hline
types & \multicolumn{10}{c|}{natural images} & \multicolumn{4}{c}{non-natural images} \\
\hline
tasks & ACT & CEL & COL & CNT & FIG & OBJ & OCR & POS & SCE & mean & CR & NC & TT & mean \\
\hline
vanilla & 73.6 & 74.0 & 73.4 & 59.8 & 82.8 & 79.6 & 77.0 & 65.8 & 87.0 & 74.8 & \textbf{27.7} & 21.0 & \textbf{2.7} & \textbf{17.1} \\
AAT-GA & \textbf{75.8} & \textbf{77.7} & \textbf{76.8} & 60.4 & 86.4 & 81.4 & \textbf{81.3} & 67.4 & \textbf{89.0} & 77.4 & 20.3 & \textbf{21.3} & 2.3 & 14.6 \\
AAT-BP & 74.8 & 77.0 & 76.2 & \textbf{60.6} & \textbf{86.6} & \textbf{83.0} & 79.7 & \textbf{71.6} & 88.0 & \textbf{77.5} & 22.3 & 18.3 & 2.3 & 14.3 \\
\hline
\end{tabular}
\vspace{-2pt}
\caption{Text-to-image retrieval results on ReCoS using ViT-B, evaluated with R@1 as the metric.}\label{tab:recos-t2i}
\vspace{-2pt}
\end{table*}

\section{Experiments}\label{sec:exp}
\subsection{Experimental Settings}\label{sec:imp}
\subsubsection{Models}
We evaluate AAT using CLIP~\cite{radford2021learning} and Chinese-CLIP~\cite{yang2022chinese}, across ViT-B, ViT-L, and ViT-H. Specifically, we use OpenCLIP released models~\cite{ilharco} trained on LAION-2B (a subset of LAION-5B~\cite{schuhmann2022laion}) and Chinese-CLIP released models trained on 200M Chinese image-text pairs. 

\subsubsection{AAT Optimization Data}\label{sec:data}
As introduced in Sections~\ref{sec:aat-ga} and~\ref{sec:aat-bp}, we use a small validation set $\mathcal{D}$ to optimize AAT. For CLIP, we randomly sample 1k image-text pairs from the MS COCO \emph{train} set as $\mathcal{D}$. For Chinese-CLIP, we directly use the COCO-CN \emph{val} set that contains 1k image-text pairs.

\subsubsection{AAT-GA Implementation}
The key hyper-parameters for GA are configured as follows: the population size is determined by the number of heads---48 for ViT-B, 96 for ViT-L, and 128 for ViT-H. A two-point crossover is applied with a probability of 0.9, while mutation occurs with a probability of 0.5 using the flip-bit strategy. The tournament selection with a size of 3 is employed. The evolution proceeds for up to 100 generations, with early stopping triggered when fitness gains plateau or population diversity becomes low.

\subsubsection{AAT-BP Implementation}
The BP optimization settings are as follows: learning rate of 2e-2 for CLIP and 5e-2 for Chinese-CLIP, with a batch size of 256. ViT-B models are trained for 32 epochs, and larger models for 64. All other hyper-parameters follow the OpenCLIP codebase~\cite{ilharco}.

\subsubsection{Benchmarks}
We evaluate AAT-improved CLIP upon MS COCO~\cite{lin2014microsoft}, Flickr30k~\cite{young2014image}, and ReCoS~\cite{chen2024recos}, and assess AAT-improved Chinese-CLIP on two widely-used Chinese retrieval benchmarks: COCO-CN~\cite{li2019coco} and Flickr30k-CNA~\cite{xie2023ccmb}. For MS COCO, Flickr30k, COCO-CN, Flickr30k-CNA, we report results on two splits: the \emph{test} split (the original test set) and the \emph{all} split (all samples in the dataset), enabling a comprehensive large-scale retrieval analysis. We further evaluate zero-shot classification on the ImageNet-1k \emph{val} set~\cite{deng2009imagenet}. Dataset and metric details are provided in Appendix~\ref{app:data}.

\subsubsection{Baseline}
Aside from vanilla CLIP, we also compare AAT to two recent strong PEFT baselines, namely CLIP-LoRA~\cite{zanella2024low} and ReFT~\cite{wu2024reft}. Detailed settings for both are as follows.

\paragraph{Vanilla CLIP} The vanilla CLIP and Chinese-CLIP models with their original, unmodified weights (prior to any improvement via AAT) serve as our primary baselines.

\paragraph{CLIP-LoRA} For fairness, we apply LoRA (using rank 1 or 2) to the Q/K projections of each head from image encoder, mirroring AAT's intervention points. To better match AAT-BP's parameter count, we also include a light-weight variant (LoRA-lite) that targets only the final image encoder layer. We adopt CLIP-LoRA's training setup: learning rate 2e-4, batch size 32, and 64 epochs (chosen via search).

\paragraph{ReFT} Following the paper, we use rank 4 for the low-rank projection $R$. A preparatory search over intervention positions shows that applying ReFT to all tokens slightly outperforms restricting it to the class token or the top-5 tokens for CLIP (contrary to some LLM findings). For completeness, we implement both DiReFT and LoReFT across all image-encoder layers, and parameter-reduced variants (DiReFT- / LoReFT-lite) applied only to the last layer. We train with learning rate 2e-5 and batch size 32 (higher rates tend to collapse $R$), for 96 epochs until accuracy saturates.

\subsection{Compared with Vanilla CLIP}
\subsubsection{Cross-modal Retrieval}\label{sec:res-clip}
\paragraph{MS COCO}
Table~\ref{tab:coco-t2i-i2t} reports MS COCO results for text-to-image and image-to-text retrieval with AAT-GA and AAT-BP. For text-to-image, GA and BP perform comparably, with mean-R differences under 1\%. On the \emph{test} set, AAT yields a 1.3\%$\sim$2.6\% gains in mean-R across model sizes, and over 1.5\% on average for the \emph{all} set. For image-to-text, AAT achieves a 0.8\%$\sim$2.1\% mean-R gain on the \emph{test} set and 0.6\%$\sim$2.7\% on the \emph{all} set.

\paragraph{Flickr30k}
Table~\ref{tab:flickr30k-t2i-i2t} reports Flickr30k results. For text-to-image, AAT presents a 1.08\%$\sim$1.93\% mean-R improvement on the \emph{test} set and 1.64\%$\sim$2.56\% on the \emph{all} set. Notably, gains on the \emph{test} set tend to diminish at higher recall levels, likely due to saturation. A similar trend can be observed for image-to-text retrieval as well, yet with consistent mean-R improvements of 0$\sim$0.8\% on the \emph{test} set and 1.08\%$\sim$3.61\% on the \emph{all} set.

\begin{table*}[ht]
\centering
\fontsize{7}{9}\selectfont
\begin{tabular}{p{10pt}<{\centering}|p{35pt}<{\centering}|p{12pt}<{\centering}p{12pt}<{\centering}p{12pt}<{\centering}p{12pt}<{\centering}|p{12pt}<{\centering}p{12pt}<{\centering}p{12pt}<{\centering}p{12pt}<{\centering}|p{12pt}<{\centering}p{12pt}<{\centering}p{12pt}<{\centering}p{12pt}<{\centering}|p{12pt}<{\centering}p{12pt}<{\centering}p{12pt}<{\centering}p{12pt}<{\centering}}
\hline
\multicolumn{2}{c|}{tasks} & \multicolumn{8}{c|}{text-to-image retrieval} & \multicolumn{8}{c}{image-to-text retrieval} \\
\hline
\multicolumn{2}{c|}{splits} & \multicolumn{4}{c|}{the \emph{test} set} & \multicolumn{4}{c|}{the \emph{all} set} & \multicolumn{4}{c|}{the \emph{test} set} & \multicolumn{4}{c}{the \emph{all} set} \\
\hline
\multicolumn{1}{c|}{size} & \multicolumn{1}{c|}{method} & R@1 & R@5 & R@10 & mR & R@1 & R@5 & R@10 & mR & R@1 & R@5 & R@10 & mR & R@1 & R@5 & R@10 & mR \\
\hline
\multirow{3}{*}{B} & vanilla & 62.2 & 86.9 & 94.3 & 81.1 & 24.0 & 45.4 & 55.8 & 41.7 & 56.2 & 83.8 & 93.4 & 77.8 & 22.6 & 43.0 & 52.7 & 39.4 \\
 & AAT-GA & 64.9 & 89.9 & \textbf{96.5} & 83.8 & \textbf{27.6} & 50.0 & 60.4 & 46.0 & 62.6 & 89.0 & 95.3 & 82.3 & 25.5 & 47.1 & 57.2 & 43.3 \\
 & AAT-BP & \textbf{66.0} & \textbf{90.7} & 96.3 & \textbf{84.3} & 27.4 & \textbf{50.4} & \textbf{60.6} & \textbf{46.1} & \textbf{63.1} & \textbf{90.4} & \textbf{95.6} & \textbf{83.0} & \textbf{26.2} & \textbf{48.3} & \textbf{58.1} & \textbf{44.2} \\
\hline
\multirow{3}{*}{L} & vanilla & 63.9 & 88.7 & 94.5 & 82.4 & 26.7 & 48.6 & 58.5 & 44.6 & 60.4 & 84.2 & 93.3 & 79.3 & 24.4 & 44.6 & 54.5 & 41.2 \\
 & AAT-GA & 66.5 & \textbf{91.4} & 96.1 & 84.7 & 30.1 & 52.9 & 63.2 & 48.7 & 65.9 & 89.0 & 95.4 & 83.4 & 29.8 & 51.6 & 61.6 & 47.7 \\
 & AAT-BP & \textbf{68.2} & 91.1 & \textbf{96.6} & \textbf{85.3} & \textbf{30.2} & \textbf{52.9} & \textbf{63.3} & \textbf{48.8} & \textbf{67.3} & \textbf{90.6} & \textbf{96.9} & \textbf{84.9} & \textbf{30.4} & \textbf{52.8} & \textbf{63.5} & \textbf{48.9} \\
\hline
\multirow{3}{*}{H} & vanilla & 69.6 & 89.9 & 95.8 & 85.1 & 30.4 & 52.4 & 62.2 & 48.3 & 63.1 & 86.7 & 93.0 & 80.9 & 26.3 & 46.5 & 56.0 & 42.9 \\
 & AAT-GA & 72.3 & 92.3 & 96.5 & 87.0 & 33.8 & 56.8 & 66.6 & 52.4 & 66.6 & 89.0 & 94.5 & 83.4 & 30.0 & 51.1 & 60.4 & 47.1 \\
 & AAT-BP & \textbf{73.7} & \textbf{92.7} & \textbf{97.6} & \textbf{88.0} & \textbf{35.5} & \textbf{59.4} & \textbf{69.1} & \textbf{54.7} & \textbf{72.4} & \textbf{93.1} & \textbf{97.3} & \textbf{87.6} & \textbf{35.4} & \textbf{58.4} & \textbf{68.3} & \textbf{54.0} \\
\hline
\end{tabular}
\vspace{-2pt}
\caption{Comparison on COCO-CN for cross-modal retrieval. The symbols in the table have the same meaning as those in Table~\ref{tab:coco-t2i-i2t}.}\label{tab:cococn-t2i-i2t}
\end{table*}

\begin{table*}[ht]
\centering
\fontsize{7}{9}\selectfont
\begin{tabular}{p{10pt}<{\centering}|p{35pt}<{\centering}|p{12pt}<{\centering}p{12pt}<{\centering}p{12pt}<{\centering}p{12pt}<{\centering}|p{12pt}<{\centering}p{12pt}<{\centering}p{12pt}<{\centering}p{12pt}<{\centering}|p{12pt}<{\centering}p{12pt}<{\centering}p{12pt}<{\centering}p{12pt}<{\centering}|p{12pt}<{\centering}p{12pt}<{\centering}p{12pt}<{\centering}p{12pt}<{\centering}}
\hline
\multicolumn{2}{c|}{tasks} & \multicolumn{8}{c|}{text-to-image retrieval} & \multicolumn{8}{c}{image-to-text retrieval} \\
\hline
\multicolumn{2}{c|}{splits} & \multicolumn{4}{c|}{the \emph{test} set} & \multicolumn{4}{c|}{the \emph{all} set} & \multicolumn{4}{c|}{the \emph{test} set} & \multicolumn{4}{c}{the \emph{all} set} \\
\hline
\multicolumn{1}{c|}{size} & \multicolumn{1}{c|}{method} & R@1 & R@5 & R@10 & mR & R@1 & R@5 & R@10 & mR & R@1 & R@5 & R@10 & mR & R@1 & R@5 & R@10 & mR \\
\hline
\multirow{3}{*}{B} & vanilla & 62.12 & 86.62 & 92.52 & 80.42 & 24.35 & 44.78 & 54.36 & 41.16 & 73.70 & 93.20 & 97.00 & 87.97 & 33.45 & 56.84 & 66.56 & 52.28 \\
 & AAT-GA & 65.60 & 89.04 & 94.18 & 82.94 & 26.79 & 48.20 & 57.91 & 44.30 & 77.70 & \textbf{95.90} & 97.50 & 90.37 & 36.76 & 60.51 & 70.10 & 55.79 \\
 & AAT-BP & \textbf{66.58} & \textbf{89.28} & \textbf{94.36} & \textbf{83.41} & \textbf{27.68} & \textbf{49.51} & \textbf{59.17} & \textbf{45.45} & \textbf{78.60} & 95.50 & \textbf{97.90} & \textbf{90.67} & \textbf{37.18} & \textbf{61.46} & \textbf{71.07} & \textbf{56.57} \\
\hline
\multirow{3}{*}{L} & vanilla & 67.58 & 89.52 & 94.28 & 83.79 & 28.79 & 50.39 & 59.80 & 46.33 & 80.00 & 96.50 & 98.20 & 91.57 & 37.15 & 61.59 & 71.05 & 56.59 \\
 & AAT-GA & \textbf{71.48} & \textbf{91.60} & 95.52 & \textbf{86.20}& \textbf{32.45} & \textbf{55.02} & \textbf{64.51} & \textbf{50.66} & 87.00 & \textbf{97.60} & 98.90 & 94.50 & \textbf{45.17} & \textbf{69.78} & \textbf{78.53} & \textbf{64.50} \\
 & AAT-BP & 70.90 & 91.34 & \textbf{95.70} & 86.00 & 31.69 & 54.46 & 64.10 & 50.08 & \textbf{87.50} & 97.40 & \textbf{99.10} & \textbf{94.67} & 44.25 & 69.05 & 77.82 & 63.71 \\
\hline
\multirow{3}{*}{H} & vanilla & 70.94 & 91.30 & 95.30 & 85.85 & 33.92 & 56.27 & 65.38 & 51.86 & 81.40 & 97.00 & 98.90 & 92.43 & 42.35 & 66.47 & 75.43 & 61.42 \\
 & AAT-GA & 74.86 & 93.00 & 96.50 & 88.12 & 37.54 & 60.46 & 69.47 & 55.82 & 85.40 & \textbf{98.10} & 99.00 & 94.17 & 48.23 & 72.16 & 80.29 & 66.89 \\
 & AAT-BP & \textbf{75.30} & \textbf{93.40} & \textbf{96.78} & \textbf{88.49} & \textbf{37.64} & \textbf{60.68} & \textbf{69.64} & \textbf{55.99} & \textbf{88.60} & 97.50 & \textbf{99.30} & \textbf{95.13} & \textbf{52.12} & \textbf{75.71} & \textbf{83.38} & \textbf{70.40} \\
\hline
\end{tabular}
\vspace{-2pt}
\caption{Comparison on Flickr30k-CNA for cross-modal retrieval. The symbols in the table have the same meaning as those in Table~\ref{tab:coco-t2i-i2t}.}\label{tab:flickr30kcna-t2i-i2t}
\end{table*}

\paragraph{ReCoS}
To evaluate broader effectiveness, Table~\ref{tab:recos-t2i} reports text-to-image retrieval on the challenging ReCoS, spanning 12 subsets across diverse domains (subset abbreviations in Appendix~\ref{app:data}). We group subsets by image domain: natural (\emph{e.g.}, landscapes, animals, movie scenes, profile photos) and non-natural (\emph{e.g.}, code snippets, characters, formulas) that are typically synthetic with uniform backgrounds.

AAT consistently outperforms on natural subsets, with a mean R@1 gain of around 3\%. However, its performance declines on non-natural images. We attribute this drop to two main factors: (1) significant domain shifts between the test images and those used to identify detrimental attention heads (see Appendix~\ref{app:fail}); and (2) the inherent limitations of CLIP models. Tasks such as CR (code reasoning), NC (numerical calculation), and TT (text translation) demand beyond visual perception, but understanding and reasoning capabilities that CLIP is not explicitly trained for.

\paragraph{COCO-CN}
Table~\ref{tab:cococn-t2i-i2t} reports the COCO-CN results. For text-to-image retrieval, AAT achieves a 1.9\%$\sim$3.2\% higher mean-R across model sizes using both GA and BP on the \emph{test} set, while the gain is even more pronounced, reaching up to 6.4\%, on the \emph{all} set. For image-to-text retrieval, AAT delivers a 2.5\%$\sim$6.7\% improvement on the \emph{test} set and 3.9\%$\sim$11.1\% on the \emph{all} set. These advances surpass those observed in CLIP, suggesting its effectiveness is correlated with the quality of the original representations.

\paragraph{Flickr30k-CNA}
Table~\ref{tab:flickr30kcna-t2i-i2t} presents the Flickr30k-CNA results. AAT improves mean-R for text-to-image retrieval by 2.21\%$\sim$3.2\% on the \emph{test} set and 3.14\%$\sim$4.33\% on the \emph{all} set. For image-to-text, it outperforms the baseline by 3\% on the \emph{test} set and up to 8.98\% on the \emph{all} set. Consistent with COCO-CN, AAT delivers greater gains for Chinese-CLIP.

\begin{table*}[ht]
\centering
\fontsize{7}{9}\selectfont
\begin{tabular}{l|p{18pt}<{\centering}p{12pt}<{\centering}p{12pt}<{\centering}p{12pt}<{\centering}p{12pt}<{\centering}p{12pt}<{\centering}|p{18pt}<{\centering}p{12pt}<{\centering}p{12pt}<{\centering}p{12pt}<{\centering}p{12pt}<{\centering}p{12pt}<{\centering}|p{18pt}<{\centering}p{12pt}<{\centering}p{12pt}<{\centering}p{12pt}<{\centering}p{12pt}<{\centering}p{12pt}<{\centering}}
\hline
 & \multicolumn{6}{c|}{ViT-B}& \multicolumn{6}{c|}{ViT-L}& \multicolumn{6}{c}{ViT-H} \\ \cline{2-19} 
 & \multirow{2}{*}{\#param} & \multicolumn{2}{c}{MS COCO} & \multicolumn{2}{c}{Flickr30k} & \multirow{2}{*}{Avg} & \multirow{2}{*}{\#param} & \multicolumn{2}{c}{MS COCO} & \multicolumn{2}{c}{Flickr30k} & \multirow{2}{*}{Avg} & \multirow{2}{*}{\#param} & \multicolumn{2}{c}{MS COCO} & \multicolumn{2}{c}{Flickr30k} & \multirow{2}{*}{Avg} \\ \cline{3-6} \cline{9-12} \cline{15-18}
 & & T2I & I2T & T2I & I2T& & & T2I & I2T & T2I & I2T & & & T2I & I2T & T2I & I2T &\\
\hline
vanilla & n/a & 58.9 & 76.0 & 82.98 & 94.30 & 78.05 & n/a & 63.7 & 77.0 & 86.91 & 94.90 & 80.63 & n/a & 66.8 & 78.7 & 88.65 & 95.63 & 82.45 \\
\hline
AAT-GA & 0 & 61.5 & 77.4 & 84.91 & \textbf{94.40} & 79.55 & 0 & 66.1 & \textbf{79.1} & 88.13 & \textbf{95.33} & \textbf{82.17} & 0 & 68.7 & 80.4 & 89.76 & 96.30 & 83.79 \\
AAT-BP & 144 & 60.9 & 77.0 & 84.60 & 94.33 & 79.21 & 384 & 66.2 & 77.8 & \textbf{88.43} & 95.00 & 81.86 & 512 & 68.1 & \textbf{80.4} & 89.73 & \textbf{96.40} & 83.66 \\
\hline
LoRA-lite & 3.1k & 59.2 & 76.3 & 83.23 & 94.10 & 78.21 & 4.1k & 64.2 & 77.4 & 87.28 & 94.60 & 80.87 & 5.1k & 67.2 & 79.0 & 88.83 & 95.60 & 82.66 \\
LoRA (r=1) & 36.9k & 62.2 & 77.2 & 85.39 & 93.63 & 79.61 & 98.3k & 65.8 & 78.7 & 88.04 & 95.27 & 81.95 & 163.8k & 68.7 & 79.9 & 89.93 & 96.10 & 83.66 \\
LoRA (r=2) & 73.8k & \textbf{62.7} & \textbf{77.5} & \textbf{85.73} & 93.98 & \textbf{79.98} & 196.6k & \textbf{66.4} & 78.6 & 88.29 & 95.22 & 82.13 & 327.5k & \textbf{69.1} & 80.3 & 89.85 & 96.40 & \textbf{83.91} \\
\hline
DiReFT-lite & 6.1k & 60.1 & 75.7 & 83.82 & 93.83 & 78.36 & 8.2k & 64.3 & 76.9 & 87.32 & 94.43 & 80.74 & 10.2k & 67.1 & 78.9 & 88.91 & 95.90 & 82.70 \\
DiReFT & 73.8k & 62.0 & 76.3 & 85.18 & 93.50 & 79.25 & 196.6k & 65.7 & 77.4 & 88.16 & 94.50 & 81.44 & 327.5k & 68.9 & 78.6 & \textbf{89.95} & 95.50 & 83.24 \\
LoReFT-lite & 6.1k & 59.9 & 75.2 & 83.49 & 93.73 & 78.08 & 8.2k & 64.3 & 77.1 & 87.11 & 94.53 & 80.76 & 10.2k & 67.0 & 78.4 & 88.67 & 94.27 & 82.09 \\
LoReFT & 73.8k & 61.1 & 75.2 & 84.70 & 93.93 & 78.73 & 196.6k & 65.4 & 76.7 & 87.76 & 94.10 & 80.99 & 327.5k & 67.8 & 78.3 & 89.47 & 95.33 & 82.73 \\
\hline
\end{tabular}
\vspace{-2pt}
\caption{Mean-R comparison for text-to-image (T2I) and image-to-text (I2T) retrieval across base, large, and huge model variants.}\label{tab:peft}
\end{table*}

\begin{table}[ht]
\centering
\fontsize{7}{9}\selectfont
\begin{tabular}{p{26pt}<{\centering}|p{10pt}<{\centering}p{10pt}<{\centering}p{12pt}<{\centering}|p{8pt}<{\centering}p{8pt}<{\centering}p{9pt}<{\centering}|p{10pt}<{\centering}p{10pt}<{\centering}p{11pt}<{\centering}}
\hline
datasets & \multicolumn{3}{c|}{ImageNet-1k} & \multicolumn{3}{c|}{MS COCO} & \multicolumn{3}{c}{Flickr30k}\\
\hline
sizes & B & L & H & B & L & H & B & L & H \\
\hline
vanilla & 68.12 & 72.62 & 76.17 & 58.9 & 63.7 & 66.8 & 82.98 & 86.91 & 88.65 \\
AAT-GA & \textbf{69.17} & \textbf{74.11} & \textbf{77.28} & \textbf{61.2} & \textbf{65.8} & \textbf{68.6} & \textbf{84.67} & 88.02 & \textbf{89.67} \\
AAT-BP & 69.01 & 73.76 & 77.25 & 61.0 & 65.6 & 68.2 & 84.42 & \textbf{88.34} & 89.61 \\
\hline
\end{tabular}
\vspace{-2pt}
\caption{Top-1 accuracy on ImageNet-1k \emph{val} set for zero-shot classification, and mean-R on MS COCO and Flickr30k \emph{test} sets.}\label{tab:zero}
\end{table}

\subsubsection{Zero-shot Classification}
Since AAT is primarily designed for retrieval, we extend its application to zero-shot classification on ImageNet-1k, with minimal but essential adaptation. We identify a mismatch between the text templates used in ImageNet-1k classification (\emph{e.g.}, ``a photo of a {CLASS\_NAME}'') and the natural language captions in the original optimization set $\mathcal{D}$ (\emph{e.g.}, ``An old woman sits on a bench''). To resolve this, we construct a new $\mathcal{D}$ by incorporating 500 image-text pairs from the ImageNet-1k training set that follow the zero-shot template style, and re-run AAT on CLIP using this adapted $\mathcal{D}$.

We test the resulting models on the ImageNet-1k \emph{val} set and observe consistent top-1 accuracy gains of 1\%$\sim$1.5\% (Table~\ref{tab:zero}). To further assess generalization, we test retrieval on MS COCO and Flickr30k. Performance remains strong, with no noticeable degradation relative to models optimized on the original $\mathcal{D}$ (Tables~\ref{tab:coco-t2i-i2t} and~\ref{tab:flickr30k-t2i-i2t}); text-to-image differences are under 0.6\% on MS COCO and 0.3\% on Flickr30k.

\subsection{Compared with PEFT}
Table~\ref{tab:peft} reports recall for text-to-image and image-to-text retrieval on MS COCO and Flickr30k, along with the number of additional learnable parameters.

\subsubsection{AAT \emph{vs.} LoRA.} In general, AAT achieves comparable recall performance to LoRA: the average mean-R difference on both MS COCO and Flickr30k is less than 0.5\%. In terms of parameter efficiency, AAT-GA adds no learnable parameters, and AAT-BP uses only hundreds, whereas LoRA requires two orders of magnitude more (roughly 250$\times$$\sim$600$\times$) than AAT-BP. When constrained to a closely comparable parameter budget (LoRA-lite, 10$\times$ AAT-BP), LoRA's performance degrades substantially. These results underscore AAT's strong parameter efficiency, making it well-suited for the resource-constrained platforms.

\subsubsection{AAT \emph{vs.} ReFT.} Overall, AAT surpasses DiReFT and LoReFT across different model sizes. In detail, AAT outperforms the DiReFT-lite and LoReFT-lite, even though these two variants require 20$\times$$\sim$40$\times$ more parameters, by 1.1$\sim$1.4 mean-R on average across base, large, and huge models. It still slightly exceeds the full ReFT variants that use about 500$\times$ more parameters. Although ReFT is often more parameter-efficient than LoRA in LLMs, these CLIP results invert this trend. Given that our LoRA setup targets Q/K projections (implicitly modifying the attention weights), we conjecture that, for compact VLMs, attention manipulation is a more effective lever than downstream representation intervention.

\subsection{Ablation Study}
\subsubsection{The Value of $\beta$}
\begin{figure}[t]
\centering
\includegraphics[width=3.2in]{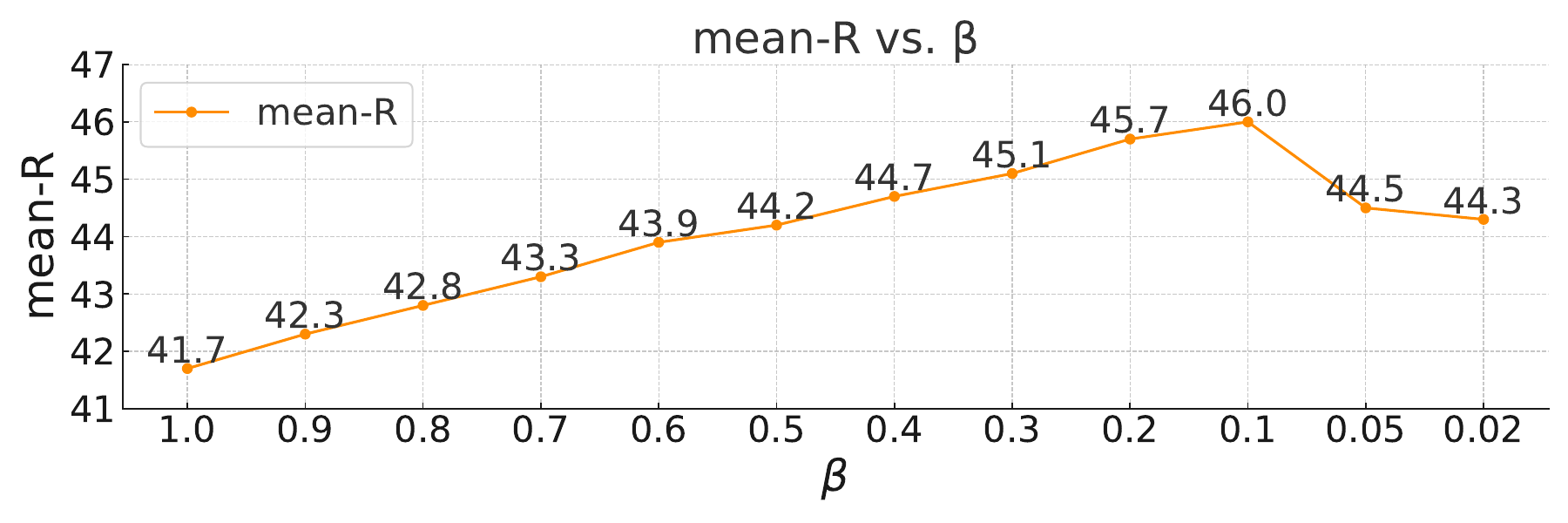}
\vspace{-5pt}
\caption{Mean-R for text-to-image retrieval on the COCO-CN \emph{all} set \emph{vs.} $\beta$ values in AAT-GA, using the ViT-B-based model.}
\vspace{-5pt}
\label{fig:beta}
\end{figure}

We empirically set $\beta=0.1$ in AAT-GA and examine its effect on performance. Figure~\ref{fig:beta} plots mean-R on the COCO-CN \emph{all} set as $\beta$ decreases from 1 to 0.02: performance rises, peaks at $\beta=0.1$, then declines.
For AAT-BP, we analyze the head-specific $\beta$ values obtained by applying a sigmoid (with temperature $\tau$) to the learnable $\alpha$ parameters. Deeper layers exhibit more polarized $\beta$ (near 0 or 1), whereas shallower layers remain closer to the initial value of 1. This suggests depth-dependent gradient propagation during BP, which may account for the result differences between AAT-GA and AAT-BP in certain cases.

\subsubsection{The Size of $D$}\label{sec:dsize}
We study the effect of the validation set size for AAT optimization. Starting from $\mathcal{D}$ with 1k image–text pairs, we randomly sample subsets of 500, 200, and 100 pairs and report mean-R for both CLIP and Chinese-CLIP in Table~\ref{tab:dsize}. Even with only 100 pairs, AAT surpasses the baseline ($\text{size}=0$) by 1\% on CLIP and 3\% on Chinese-CLIP. As increasing the size to 500, it yields only marginal additional gains.

\subsubsection{Selection of AAT Optimization Data}
As described in Section~\ref{sec:data}, we optimize AAT on a 1k subset from MS COCO or COCO-CN. To examine the effect of used data and potential domain bias, we construct an in-domain $\mathcal{D}$ using 1k Flickr30k-CNA \emph{val} image-text pairs. We then re-run AAT with this in-domain set and compare Flickr30k-CNA performance to AAT optimized on COCO-CN. As shown in Table~\ref{tab:flickr30kcna-t2i-data}, using in-domain data yields only marginal gains (typically $<0.5\%$), indicating strong cross-domain generalization of AAT-refined representations.

\subsection{AAT Optimization Runtime}\label{sec:runtime}
We optimize AAT on $8{\times}$ NVIDIA RTX 4090 GPUs with an Intel(R) Xeon(R) Platinum 8358P @ 2.60GHz platform. To demonstrate real-world feasibility, we list wall-clock runtime and GPU-hours under the standard setup in Section~\ref{sec:imp} (see the 4th–5th columns of Table~\ref{tab:cost}). While GA is attractive for low-precision (INT8/INT16) inference hardware, as requiring no gradients, it is roughly $20{\times}$ slower than BP due to a vast number of fitness evaluations, even with a small $\mathcal{D}$.

AAT is designed as an ``out-of-the-box'' solution where GA hyper-parameters (\emph{e.g.}, generations, crossover/mutation rates) are kept fixed across model sizes to minimize manual adjustment. This simplicity introduces redundant overhead. We therefore apply the following practical optimizations:
\begin{enumerate}
\item Skipping: Based on the analysis in Appendix~\ref{app:analysis}, where effective ablation ratios typically fall between 30\% and 40\%, we skip candidates whose ratios exceed 60\%.
\item Reducing population size for ViT-L/H: We find that the reduced population sizes maintain comparable accuracy while significantly cutting runtime.
\item Fewer generations: We find that performance plateaus in the final 30\% of generations, despite a continued fitness gain, suggesting that over-optimization on $\mathcal{D}$ offers diminishing benefit on test data.
\end{enumerate}

These adjustments yield substantial overhead savings---reducing GA runtime by  40\%, 70\%, and 72\% for the base, large, and huge models, with only minor performance drops (Table~\ref{tab:cost}; detailed comparisons in Appendix~\ref{app:cost}). As post-training technique, both AAT-BP and AAT-GA impose minimal cost compared to training CLIP from scratch, underscoring their practicality for real-world deployment.

\begin{table}[t]
\centering
\fontsize{7}{9}\selectfont
\begin{tabular}{p{35pt}<{\centering}p{27pt}<{\centering}p{14pt}<{\centering}p{14pt}<{\centering}p{14pt}<{\centering}p{14pt}<{\centering}p{15pt}<{\centering}}
\hline
\multicolumn{2}{c}{the size of $\mathcal{D}$} & ~~0 & ~~100 & ~~200 & ~~500 & ~1000 \\
\hline
\multicolumn{1}{c}{\multirow{2}{*}{~~~CLIP~~~}} & AAT-GA & ~~24.4 & ~~25.4 & ~~25.7 & ~~26.0 & ~~26.1 \\
\cline{2-2}
\multicolumn{1}{c}{} & AAT-BP & ~~24.4 & ~~25.2 & ~~25.4 & ~~25.7 & ~~25.9 \\
\hline
\multicolumn{1}{c}{\multirow{2}{*}{~~~Chinese-CLIP~~~}} & AAT-GA & ~~41.7 & ~~45.1 & ~~45.3 & ~~45.9 & ~~46.0 \\
\cline{2-2}
\multicolumn{1}{c}{} & AAT-BP & ~~41.7 & ~~44.7 & ~~45.2 & ~~45.7 & ~~46.1 \\
\hline
\end{tabular}
\vspace{-2pt}
\caption{Mean-R \emph{vs.} the size of $\mathcal{D}$: CLIP is evaluated on the MS COCO \emph{all} set, and Chinese-CLIP on the COCO-CN \emph{all} set, both using ViT-B-based models, for text-to-image retrieval.}\label{tab:dsize}
\end{table}

\begin{table}[t]
\centering
\fontsize{7}{9}\selectfont
\begin{tabular}{p{9pt}<{\centering}|p{16pt}<{\centering}|p{11pt}<{\centering}p{11pt}<{\centering}p{11pt}<{\centering}p{12pt}<{\centering}|p{11pt}<{\centering}p{11pt}<{\centering}p{11pt}<{\centering}p{11pt}<{\centering}}
\hline
\multicolumn{2}{c|}{splits} & \multicolumn{4}{c|}{the \emph{test} set} & \multicolumn{4}{c}{the \emph{all} set} \\
\hline
\multicolumn{2}{c|}{metric} & R@1 & R@5 & R@10 & mR & R@1 & R@5 & R@10 & mR \\
\hline
\multicolumn{2}{c|}{vanilla} & 62.12 & 86.62 & 92.52 & 80.42 & 24.35 & 44.78 & 54.36 & 41.16 \\
\hline
\multirow{2}{*}{GA} & CO & 65.60 & 89.04 & 94.18 & 82.94 & 26.79 & 48.20 & 57.91 & 44.30 \\
 & F30k & 66.00 & 89.40 & 94.44 & 83.28 & 27.59 & 49.16 & 58.79 & 45.18 \\
\hline
\multirow{2}{*}{BP} & CO & 66.58 & 89.28 & 94.36 & 83.41 & 27.68 & 49.51 & 59.17 & 45.45 \\
 & F30k & 67.18 & 89.92 & 94.56 & 83.89 & 27.99 & 49.81 & 59.49 & 45.76 \\
\hline
\end{tabular}
\vspace{-2pt}
\caption{Flickr30k-CNA text-to-image retrieval results based on ViT-B. CO indicates that AAT is optimized on the COCO-CN \emph{val} set, while F30k indicates optimization on a 1k-sample subset of the Flickr30k-CNA \emph{val} set.}\label{tab:flickr30kcna-t2i-data}
\end{table}

\begin{table}[t]
\centering
\fontsize{7}{9}\selectfont
\begin{tabular}{p{12pt}<{\centering}|p{9pt}<{\centering}p{9pt}<{\centering}p{9pt}<{\centering}p{9pt}<{\centering}p{9pt}<{\centering}p{9pt}<{\centering}p{9pt}<{\centering}p{9pt}<{\centering}p{9pt}<{\centering}p{10pt}<{\centering}}
\hline
tasks & ACT & CEL & COL & CNT & FIG & OBJ & OCR & POS & SCE & mean \\
\hline
$\mathcal{S}_{\text{gain}}$ & 2.26 & 2.91 & 2.02 & 2.59 & 2.29 & 2.84 & 3.79 & 2.34 & 2.40 & 2.60 \\
$\mathcal{S}_{\text{loss}}$ & 2.60 & 3.87 & 2.74 & 2.77 & 2.72 & 3.05 & 4.03 & 2.74 & 3.25 & 3.09 \\
\hline
\end{tabular}
\vspace{-2pt}
\caption{Average text-related importance of ablated heads in the last 4 layers on natural ReCoS (reported values in units of 1e-2).}\label{tab:interp}
\vspace{-2pt}
\end{table}

\subsection{An Interpretability Perspective on AAT}
We investigate the interpretability of the ablated heads using TextSpan~\cite{gandelsmaninterpreting} to correlate head-level effects to linguistic semantics. Using a ViT-B-based CLIP model, for each image–text pair we extract the text and image embeddings and, following TextSpan, decompose the image embedding into 144 (12$\times$12) head-specific image embeddings. Each one is then multiplied by the text embedding to yield a 12$\times$12 matrix of text-related importance, where larger values indicate higher semantic relevance. This matrix reveals which heads in vanilla CLIP are most (or least) aligned with the text.

We probe these semantics on the natural subsets of ReCoS. Specifically, we partition all natural pairs into those that flip from incorrect to correct after applying AAT-GA ($\mathcal{S}_{\text{gain}}$) and those that flip from correct to incorrect ($\mathcal{S}_{\text{loss}}$). For each pair, we compute the cumulative text-related importance (from the 12$\times$12 matrix) of the heads ablated by AAT in the last four layers, as recommended by TextSpan. As shown in Table~\ref{tab:interp}, the ablated heads in $\mathcal{S}_{\text{gain}}$ have lower average importance, whereas those in $\mathcal{S}_{\text{loss}}$ have higher importance. This supports the view that suppressing less text-relevant heads tends to improve image–text alignment (and suppressing highly text-relevant heads can harm it)~\cite{gandelsmaninterpreting}.

These findings complement the CLIP interpretability research~\cite{gandelsmaninterpreting}, which manually removes heads associated with spurious cues in specific layers and observes improved representations. Instead, AAT offers an automated, data-driven scheme that can target detrimental heads across all layers. Encouragingly, AAT naturally focuses on heads with lower text-related importance, echoing those manual intervention results. It thereby reinforces the conclusion that attention heads exhibit meaningful semantic structure.

\begin{table}[t]
\centering
\fontsize{8}{10}\selectfont
\begin{tabular}{p{8pt}<{\centering}p{20pt}<{\centering}p{25pt}<{\centering}p{35pt}<{\centering}p{35pt}<{\centering}p{40pt}<{\centering}}
\hline
size & metric & vanilla & BP & GA & GA (opt) \\
\hline
\multirow{3}{*}{B} & mR & 58.9 & 60.9 & 61.5 & 61.2 \\
\cline{2-2}
 & runtime & - & \makecell[c]{2 min\\(0.3 GHs)} & \makecell[c]{30 min\\(4 GHs)} & \makecell[c]{18 min\\(2.4 GHs)} \\
\hline
\multirow{3}{*}{L} & mR & 63.7 & 66.2 & 66.1 & 65.9 \\
\cline{2-2}
 & runtime & - & \makecell[c]{5 min\\(0.7 GHs)} & \makecell[c]{1.5 h\\(12 GHs)} & \makecell[c]{27 min\\(3.6 GHs)} \\
\hline
\multirow{3}{*}{H} & mR & 66.8 & 68.1 & 68.7 & 68.3 \\
\cline{2-2}
 & runtime & - & \makecell[c]{8 min\\(1.1 GHs)} & \makecell[c]{3 h\\(24 GHs)} & \makecell[c]{50 min\\(6.7 GHs)} \\
\hline
\end{tabular}
\vspace{-2pt}
\caption{Mean-R on MS COCO \emph{test} set for text-to-image retrieval \emph{vs.} actual optimization runtime. ``Opt'' denotes AAT-GA with optimized hyper-parameter settings; ``GHs'' denotes the GPU hours.}\label{tab:cost}
\vspace{-5pt}
\end{table}

\section{Conclusion}
To better understand CLIP's attention mechanism, we conduct comprehensive analysis of attention heads in its image encoder. Observing heterogeneous head impacts, we propose AAT, a simple but effective method that refines representations by systematically ablating detrimental heads via attention-weight manipulation. For detrimental head identification, we present two alternative strategies: AAT-GA that suited to limited floating-point compute (\emph{e.g.}, INT8/INT16 edge devices), and AAT-BP which offers greater optimization efficiency when resources allow. Experimental results demonstrate that AAT improves CLIP-family models across diverse downstream tasks, achieves far greater parameter efficiency than PEFT, and corroborates recent interpretability findings. We hope this study contributes to deeper understanding of CLIP and paves the way to efficiency of VLMs.

{
    \small
    \bibliographystyle{ieeenat_fullname}
    \bibliography{main}
}


\clearpage
\setcounter{page}{1}
\maketitlesupplementary

\section{Benchmark Details and Evaluation Metrics}\label{app:data}
\subsection{MS COCO}
MS COCO is a large-scale dataset with 91 object categories in natural contexts, comprising about 113,287 training, 5,000 validation, and 5,000 test images (with annotations not publicly available), annotated with bounding boxes, segmentation, and 5 descriptive texts per image. For our experiments, we reconstruct the available training and validation images into two test sets: the \emph{test} set, which corresponds to the original validation set, and the \emph{all} set, which includes both the training and validation images. We evaluate MS COCO for text-to-image and image-to-text retrieval using recall rates at k (R@1, R@5, R@10) and mean-R (the average of R@1, R@5, and R@10) as the evaluation metrics.

\subsection{Flickr30k}
Flickr30k contains 31k images collected from Flickr, each accompanied by 5 reference sentences provided by human annotators. The dataset is split into 29k training images, 1k validation images, and 1k test images. Similarly to MS COCO, we treat the test images as the \emph{test} set and the entire dataset as the \emph{all} set. Both reconstructed sets are used to evaluate our method for text-to-image and image-to-text retrieval, with recall rates as the evaluation metric.

\subsection{ReCoS}
ReCoS is a benchmark designed for image-text retrieval in real-world scenarios, comprising 12 overlapping subsets with a total of 1k test images and diverse annotations per subset. We follow the standardized ReCoS-v1 evaluation protocol for experiments and report R@1 performance for text-to-image retrieval across all 12 subsets, as done in the original ReCoS paper. The subset abbreviations in Table~\ref{tab:recos-t2i} are: ACT for \emph{action}, CEL for \emph{celebrity}, COL for \emph{color}, CNT for \emph{count}, FIG for \emph{figure}, OBJ for \emph{object}, POS for \emph{position}, SCE for \emph{scene}, CR for \emph{code reasoning}, NC for \emph{numerical calculation}, and TT for \emph{text translation}.

\subsection{COCO-CN}
COCO-CN is a Chinese image-text retrieval dataset with about 18k training, 1k validation, and 1k test images, annotated with 27,218 manually written Chinese sentences. In our experiments, we use the COCO-CN test images as the \emph{test} set and all images as the \emph{all} set, evaluating both sets with the same metric used for MS COCO.

\subsection{Flickr30k-CNA}
Flickr30k-CNA is a cleaned Chinese-translated version of Flickr30k, with captions generated by professional English–Chinese linguists. We process and evaluate it following the same protocol used for Flickr30k.

\subsection{ImageNet-1k}
ImageNet-1k is a widely used image classification benchmark originally curated for the Large Scale Visual Recognition Challenge (ILSVRC). It contains 1k object categories, with over 1.2 million labeled training images, 50,000 validation images, and 100,000 test images. ImageNet-1k also serves as a standard benchmark for zero-shot classification. During inference, object class names (provided in the CLIP repository\footnote{\url{https://github.com/OFA-Sys/Chinese-CLIP/blob/master/zeroshot_dataset_en.md}}) are converted into textual prompts using several predefined templates\footnote{\url{https://github.com/LAION-AI/CLIP_benchmark/blob/main/clip_benchmark/datasets/en_zeroshot_classification_templates.json}}. In our experiments, we evaluate zero-shot classification performance on the ImageNet-1k \emph{val} set using top-1 accuracy as the metric.

\subsection{\emph{Cola}}
\emph{Cola}~\cite{ray2024cola} is a compositional text-to-image retrieval benchmark designed to evaluate a model's ability to localize and compose objects with their corresponding attributes. It includes both real-world images and synthetic 3D-rendered scenes. The task requires retrieving images that match the correct configuration of objects and attributes while rejecting distractors that contain the right components in incorrect arrangements. \emph{Cola} consists of approximately 1.2k composed queries involving 168 objects and 197 attributes, distributed across roughly 30K images. It features two types of queries: single-object compositions (sourced from GQA, CLEVR, and PACO) and multi-object queries. Following the original paper, mean average precision (mAP) is used for evaluating single-object subsets, while mean accuracy is used for the multi-object subset.

\begin{table*}[ht]
\centering
\small
\begin{tabular}{p{13pt}<{\centering}|p{40pt}<{\centering}|p{13pt}<{\centering}p{13pt}<{\centering}p{16pt}<{\centering}p{13pt}<{\centering}|p{13pt}<{\centering}p{13pt}<{\centering}p{16pt}<{\centering}p{13pt}<{\centering}|p{13pt}<{\centering}p{13pt}<{\centering}p{16pt}<{\centering}p{13pt}<{\centering}|p{13pt}<{\centering}p{13pt}<{\centering}p{16pt}<{\centering}p{13pt}<{\centering}}
\hline
\multicolumn{2}{c|}{tasks} & \multicolumn{8}{c|}{text-to-image retrieval} & \multicolumn{8}{c}{image-to-text retrieval} \\
\hline
\multicolumn{2}{c|}{splits} & \multicolumn{4}{c|}{the \emph{test} set} & \multicolumn{4}{c|}{the \emph{all} set} & \multicolumn{4}{c|}{the \emph{test} set} & \multicolumn{4}{c}{the \emph{all} set} \\
\hline
\multicolumn{1}{c|}{size} & \multicolumn{1}{c|}{method} & R@1 & R@5 & R@10 & mR & R@1 & R@5 & R@10 & mR & R@1 & R@5 & R@10 & mR & R@1 & R@5 & R@10 & mR \\
\hline
\multirow{2}{*}{B} & vanilla & 62.2 & 86.9 & 94.3 & 81.1 & 24.0 & 45.4 & 55.8 & 41.7 & 56.2 & 83.8 & 93.4 & 77.8 & 22.6 & 43.0 & 52.7 & 39.4 \\
& naive h-a & 62.7 & 89.6 & 95.6 & 82.6 & 25.2 & 47.0 & 57.5 & 43.2 & 55.8 & 85.5 & 93.4 & 78.2 & 23.6 & 45.0 & 55.0 & 41.2 \\
\hline
\multirow{2}{*}{L} & vanilla & 63.9 & 88.7 & 94.5 & 82.4 & 26.7 & 48.6 & 58.5 & 44.6 & 60.4 & 84.2 & 93.3 & 79.3 & 24.4 & 44.6 & 54.5 & 41.2 \\
& naive h-a & 65.9 & 89.9 & 95.6 & 83.8 & 28.9 & 51.2 & 61.4 & 47.2 & 61.6 & 86.1 & 93.8 & 80.5 & 27.7 & 47.9 & 59.2 & 44.9 \\
\hline
\multirow{2}{*}{H} & vanilla & 69.6 & 89.9 & 95.8 & 85.1 & 30.4 & 52.4 & 62.2 & 48.3 & 63.1 & 86.7 & 93.0 & 80.9 & 26.3 & 46.5 & 56.0 & 42.9 \\
& naive h-a & 71.8 & 91.3 & 97.0 & 86.7 & 32.6 & 55.0 & 64.9 & 50.8 & 64.2 & 87.7 & 94.2 & 82.0 & 28.5 & 48.7 & 59.4 & 45.5 \\
\hline
\end{tabular}
\caption{Comparison on COCO-CN for retrieval. R@k denotes recall at rank k, and mR (mean-R) is the average of R@1, R@5, and R@10. Model sizes are denoted as B (ViT-B), L (ViT-L), and H (ViT-H). ``Vanilla'' refers to the baseline model, while ``naive h-a'' indicates the model enhanced via naive joint head ablation.}\label{app:tab:cococn-t2i-i2t}
\end{table*}

\section{Impact of Individual Attention Heads}\label{app:greedsearch}
As described in Section~\ref{sec:gridsearch}, we investigate the impact of each individual attention head in CLIP's image encoder on text-to-image retrieval performance. Detailed results are provided in Table~\ref{app:tab:head}. As shown, ablating even a single attention head can outperform the vanilla baseline, reaching up to 81.1\% mean-R. Conversely, removing certain critical heads leads to a significant performance drop of up to 3.7\% compared to the baseline. These results highlight the distinct roles of individual heads and suggest that carefully selecting which heads to ablate could yield further performance gains.

In the main content of the paper, we identify all negatively impactful heads---those that individually yield higher retrieval performance than the vanilla baseline when ablated---and jointly ablate them in a simple strategy referred to as naive joint head ablation (naive h-a). We extend this experiment across different model sizes for both text-to-image and image-to-text retrieval on COCO-CN. As shown in Table~\ref{app:tab:cococn-t2i-i2t}, even this straightforward method surpasses the baseline by 1.4\%$\sim$1.6\% on the \emph{test} set and 1.5\%$\sim$2.6\% on the \emph{all} set for text-to-image retrieval, and by 0.4\%$\sim$1.2\% on the \emph{test} set and 1.8\%$\sim$3.7\% on the \emph{all} set for image-to-text retrieval.

\section{Analysis of the Ablated Heads}\label{app:analysis}
\subsection{Layer Distribution}
To illustrate the effect of AAT, we present the distribution of ablated heads in AAT-GA across layers for ViT-B, ViT-L, and ViT-H-based models in Figure~\ref{fig:all-vit}, covering both English and Chinese versions of CLIP. The results reveal that CLIP models tend to have greater head redundancy (\emph{i.e.}, more detrimental heads) in the shallow and deep layers, with fewer in the intermediate layers. In contrast, Chinese-CLIP models exhibit a more balanced ablation pattern across all layers.

\begin{figure*}[htbp]
    \centering
    \makebox[\textwidth][c]{%
        \begin{minipage}[t]{0.225\textwidth}
            \centering
            \begin{tikzpicture}
            \begin{axis}[
                ybar,
                bar width=0.5pt,
                width=1.25\textwidth,
                height=4.9cm,
                enlargelimits=0.1,
                ymin=0,
                ymax=12,
                xlabel={layers},
                ylabel={num},
                ylabel style={at={(axis description cs:0.3,0.5)}, anchor=south},
                xtick=data,
                xticklabels={0,1,...,11},
                legend style={at={(0.5,0.98)}, anchor=north, draw=none, font=\scriptsize},
                symbolic x coords={0,1,2,3,4,5,6,7,8,9,10,11},
                nodes near coords,
                every node near coord/.append style={font=\fontsize{6pt}{4.8pt}\selectfont},
                tick label style={font=\scriptsize},
                ymajorgrids=true,
                grid style=dashed,
            ]
            \addplot+[style={fill=blue!60}] coordinates {
                (0,3) (1,7) (2,5) (3,4) (4,3) (5,3)
                (6,0) (7,3) (8,4) (9,5) (10,6) (11,1)
            };
            \addplot+[style={fill=orange!70}] coordinates {
                (0,2) (1,4) (2,4) (3,3) (4,3) (5,2)
                (6,5) (7,3) (8,1) (9,4) (10,7) (11,3)
            };
            \legend{CLIP,~Chinese-CLIP}
            \end{axis}
            \end{tikzpicture}
            \makebox[\linewidth]{\small ~~~(a) ViT-B-based AAT}
        \end{minipage}
        \begin{minipage}[t]{0.32\textwidth}
            \centering
            \begin{tikzpicture}
            \begin{axis}[
                ybar,
                bar width=0.5pt,
                width=1.13\textwidth,
                height=4.9cm,
                enlargelimits=0.1,
                ymin=0,
                ymax=16,
                xlabel={layers},
                ylabel={num},
                ylabel style={at={(axis description cs:0.2,0.5)}, anchor=south},
                xtick={0,2,4,6,8,10,12,14,16,18,20,22},
                xticklabels={0,2,4,6,8,10,12,14,16,18,20,22},
                legend style={at={(0.5,0.98)}, anchor=north, draw=none, font=\scriptsize},
                symbolic x coords={0,1,2,3,4,5,6,7,8,9,10,11,12,13,14,15,16,17,18,19,20,21,22,23},
                nodes near coords,
                every node near coord/.append style={font=\fontsize{6pt}{4.8pt}\selectfont},
                tick label style={font=\scriptsize},
                ymajorgrids=true,
                grid style=dashed,
            ]
            \addplot+[style={fill=blue!60}] coordinates {
                (0,5) (1,8) (2,6) (3,4) (4,5) (5,5)
                (6,4) (7,5) (8,5) (9,7) (10,5) (11,1)
                (12,1) (13,4) (14,5) (15,6) (16,7) (17,5)
                (18,0) (19,12) (20,10) (21,8) (22,6) (23,6)
            };
            \addplot+[style={fill=orange!70}] coordinates {
                (0,9) (1,7) (2,6) (3,5) (4,3) (5,3)
                (6,8) (7,5) (8,5) (9,5) (10,10) (11,6)
                (12,8) (13,6) (14,6) (15,5) (16,7) (17,6)
                (18,6) (19,7) (20,6) (21,10) (22,6) (23,4)
            };
            \legend{CLIP,~Chinese-CLIP}
            \end{axis}
            \end{tikzpicture}
            \makebox[\linewidth]{\small ~~~~(b) ViT-L-based AAT}
        \end{minipage}
        \begin{minipage}[t]{0.42\textwidth}
            \centering
            \begin{tikzpicture}
            \begin{axis}[
                ybar,
                bar width=0.1pt,
                width=1.15\textwidth,
                height=4.9cm,
                enlargelimits=0.1,
                ymin=0,
                ymax=16,
                xlabel={layers},
                ylabel={num},
                ylabel style={at={(axis description cs:0.15,0.5)}, anchor=south},
                xtick={0,2,4,6,8,10,12,14,16,18,20,22,24,26,28,30},
                xticklabels={0,2,4,6,8,10,12,14,16,18,20,22,24,26,28,30},
                legend style={at={(0.5,0.98)}, anchor=north, draw=none, font=\scriptsize},
                symbolic x coords={0,1,2,3,4,5,6,7,8,9,10,11,12,13,14,15,16,17,18,19,20,21,22,23,24,25,26,27,28,29,30,31},
                nodes near coords,
                every node near coord/.append style={font=\fontsize{6pt}{4.8pt}\selectfont},
                tick label style={font=\scriptsize},
                ymajorgrids=true,
                grid style=dashed,
            ]
            \addplot+[style={fill=blue!60}] coordinates {
                (0,7) (1,10) (2,8) (3,8) (4,5) (5,5)
                (6,4) (7,9) (8,5) (9,7) (10,5) (11,2)
                (12,7) (13,5) (14,1) (15,8) (16,9) (17,7)
                (18,7) (19,7) (20,9) (21,7) (22,4) (23,8)
                (24,8) (25,3) (26,5) (27,4) (28,9) (29,5) (30,9) (31,8)
            };
            \addplot+[style={fill=orange!70}] coordinates {
                (0,10) (1,10) (2,6) (3,9) (4,12) (5,5)
                (6,8) (7,4) (8,4) (9,9) (10,7) (11,6)
                (12,1) (13,8) (14,3) (15,4) (16,8) (17,5)
                (18,5) (19,6) (20,6) (21,7) (22,8) (23,8)
                (24,6) (25,4) (26,6) (27,7) (28,4) (29,4) (30,8) (31,6)
            };
            \legend{CLIP,~Chinese-CLIP}
            \end{axis}
            \end{tikzpicture}
            \makebox[\linewidth]{\small ~~~~~~~(c) ViT-H-based AAT}
        \end{minipage}
    }
    \caption{Number of ablated heads across layers in AAT for ViT-B, ViT-L, and ViT-H. ViT-B consists of 12 layers with 12 attention heads each; ViT-L has 24 layers with 16 heads per layer; and ViT-H includes 32 layers, also with 16 heads per layer.}
    \label{fig:all-vit}
\end{figure*}

\subsection{Statistics}
We report the overall ablation ratio (calculated as the number of ablated heads divided by the total number of heads) and the average number of ablated heads per layer for different model sizes in Table~\ref{tab:stat}. As shown, both CLIP and Chinese-CLIP models follow similar trends: base models have ablation ratios around 0.3, while larger models reach up to 0.4. This supports the intuition that larger models contain more redundant attention heads.

\begin{table}[t]
\centering
\small
\begin{tabular}{p{55pt}<{\centering}p{30pt}<{\centering}p{28pt}<{\centering}p{28pt}<{\centering}p{28pt}<{\centering}}
\hline
model & metric & ViT-B & ViT-L & ViT-H \\
\hline
\multirow{2}{*}{CLIP} & ratio & 0.31 & 0.34 & 0.40 \\
\cline{2-2}
 & avg  & 3.7 & 5.4 & 6.4 \\
\hline
\multirow{2}{*}{Chinese-CLIP} & ratio & 0.28 & 0.38 & 0.40 \\
\cline{2-2}
 & avg & 3.4 & 6.1 & 6.4 \\
\hline
\end{tabular}
\caption{Overall ablation ratio and average ablated heads per layer across different model sizes and languages. The results are averaged over three AAT-GA runs with different random seeds.}\label{tab:stat}
\end{table}

\section{Computational Cost Analysis}\label{app:cost}
We optimize AAT-GA by eliminating redundant overhead, as described in Section~\ref{sec:runtime} of the paper. Detailed performance and runtime comparisons under different hyper-parameter configurations are provided in Tables~\ref{tab:cost_opt1} and~\ref{tab:cost_opt2}. 

Table~\ref{tab:cost_opt1} reports AAT-GA performance and optimization runtime across various hyper-parameter settings. Based on these results, we identify efficient GA configurations: \texttt{gen} = 75 and \texttt{psize} = 48 for ViT-B/L models, and \texttt{gen} = 75 and \texttt{psize} = 64 for ViT-H models. As shown in Table~\ref{tab:cost_opt2}, skipping fitness evaluations for high-ablation-ratio candidates further improves efficiency compared to the unified configuration. The final runtime cost and corresponding performance are summarized in Table~\ref{tab:cost} in the main paper.

\begin{table*}[ht]
\centering
\small
\begin{tabular}{p{18pt}<{\centering}|p{16pt}<{\centering}|p{9pt}<{\centering}|p{16pt}<{\centering}|p{9pt}<{\centering}|p{16pt}<{\centering}|p{9pt}<{\centering}||p{16pt}<{\centering}|p{9pt}<{\centering}|p{16pt}<{\centering}|p{9pt}<{\centering}|p{16pt}<{\centering}|p{9pt}<{\centering}||p{16pt}<{\centering}|p{9pt}<{\centering}|p{16pt}<{\centering}|p{9pt}<{\centering}|p{16pt}<{\centering}|p{9pt}<{\centering}}
\hline
\multicolumn{1}{l|}{} & \multicolumn{6}{c||}{CLIP-ViT-B} & \multicolumn{6}{c||}{CLIP-ViT-L} & \multicolumn{6}{c}{CLIP-ViT-H}                                                                      \\ \hline
psize & \multicolumn{2}{c|}{48} & \multicolumn{2}{c|}{32} & \multicolumn{2}{c||}{16} & \multicolumn{2}{c|}{96} & \multicolumn{2}{c|}{72} & \multicolumn{2}{c||}{48} & \multicolumn{2}{c|}{128} & \multicolumn{2}{c|}{96} & \multicolumn{2}{c}{64} \\
\hline
gen & \multicolumn{1}{c|}{mR} & \multicolumn{1}{c|}{T} & \multicolumn{1}{c|}{mR} & \multicolumn{1}{c|}{T} & \multicolumn{1}{c|}{mR} & T & \multicolumn{1}{c|}{mR} & \multicolumn{1}{c|}{T} & \multicolumn{1}{c|}{mR} & \multicolumn{1}{c|}{T} & \multicolumn{1}{c|}{mR} & T & \multicolumn{1}{c|}{mR} & \multicolumn{1}{c|}{T} & \multicolumn{1}{c|}{mR} & \multicolumn{1}{c|}{T} & \multicolumn{1}{c|}{mR} & T \\
\hline
100 & 61.5 & \multicolumn{1}{c|}{30} & 61.1 & \multicolumn{1}{c|}{20} & 60.4 & 10 & 66.1 & \multicolumn{1}{c|}{90} & 66.1 & \multicolumn{1}{c|}{66} & 66.0 & 45 & 68.7 & \multicolumn{1}{c|}{180} & 68.8 & \multicolumn{1}{c|}{140} & 68.6 & 90 \\
75 & 61.4 & \multicolumn{1}{c|}{22} & 61.1 & \multicolumn{1}{c|}{15} & 60.1 & 8 & 66.0 & \multicolumn{1}{c|}{66} & 65.8 & \multicolumn{1}{c|}{50} & 65.9 & 33 & 68.5 & \multicolumn{1}{c|}{140} & 140 & \multicolumn{1}{c|}{100} & 68.4 & 66 \\
50 & 60.9 & \multicolumn{1}{c|}{15} & 60.6 & \multicolumn{1}{c|}{10} & 59.9 & 5 & 65.1 & \multicolumn{1}{c|}{45} & 64.9 & \multicolumn{1}{c|}{35} & 64.5 & 24 & 68.1 & \multicolumn{1}{c|}{90} & 67.9 & \multicolumn{1}{c|}{66} & 67.6 & 45 \\
\hline
\end{tabular}
\caption{Mean-R (mR) across CLIP model sizes for text-to-image retrieval on the MS COCO \emph{test} set, along with the corresponding GA optimization runtime (T, in minutes). For clarity, ``gen'' refers to the number of generations, and ``psize'' denotes the population size.}\label{tab:cost_opt1}
\end{table*}

\begin{table*}[!ht]
\centering
\small
\begin{tabular}{p{65pt}<{\centering}|p{55pt}<{\centering}|p{55pt}<{\centering}|p{55pt}<{\centering}|p{55pt}<{\centering}|p{55pt}<{\centering}|p{55pt}<{\centering}}
\hline
 & \multicolumn{2}{c|}{CLIP-ViT-B} & \multicolumn{2}{c|}{CLIP-ViT-L} & \multicolumn{2}{c}{CLIP-ViT-H} \\
\hline
 & mean-R & runtime & mean-R & runtime & mean-R & runtime \\
\hline
non-skip & 61.4 & 22 min & 65.9 & 33 min & 68.4 & 1.1 h \\
skip & 61.2 & 18 min & 65.9 & 27 min & 68.3 & 50 min \\
\hline
\end{tabular}
\caption{Comparison of mean-R and runtime with and without skipping fitness evaluations. Models are evaluated on the MS COCO \emph{test} set for text-to-image retrieval.}\label{tab:cost_opt2}
\end{table*}

\section{Evaluation on Compositional Retrieval}\label{app:com}
Compositional retrieval is a challenging task that goes beyond conventional cross-modal retrieval by requiring a deeper understanding of object relationships, attributes, and reasoning---rather than relying solely on visual perception. In order to explore the potential of AAT in this setting, we evaluate AAT-improved CLIP on \emph{Cola}, a recently proposed challenging benchmark designed for compositional text-to-image retrieval across both real-world and synthetic domains.

Table~\ref{tab:cola} reports the results for both vanilla and AAT-improved CLIP models. AAT improves mean accuracy on the multi-object benchmark by 2.38\% with GA and 3.80\% with BP. For the single-object benchmark, AAT yields mAP gains of up to 2.27\% on the \emph{Cola}-GQA subset and up to 1.95\% on the \emph{Cola}-PACO subset. However, no improvement is observed on the \emph{Cola}-CLEVR subset. This limitation likely stems from the synthetic nature of CLEVR, which consists of 3D-rendered objects and requires complex reasoning capabilities that exceed CLIP's representational capacity. As a result, CLIP---whether enhanced with AAT or not---struggles on this subset, highlighting a boundary of current vision-language models in handling highly abstract compositional reasoning.

\section{Comparison with SFT}\label{app:sft}
\begin{figure*}[htbp]
  \centering
  \begin{subfigure}[t]{0.47\textwidth}
    \centering
    \includegraphics[width=\linewidth]{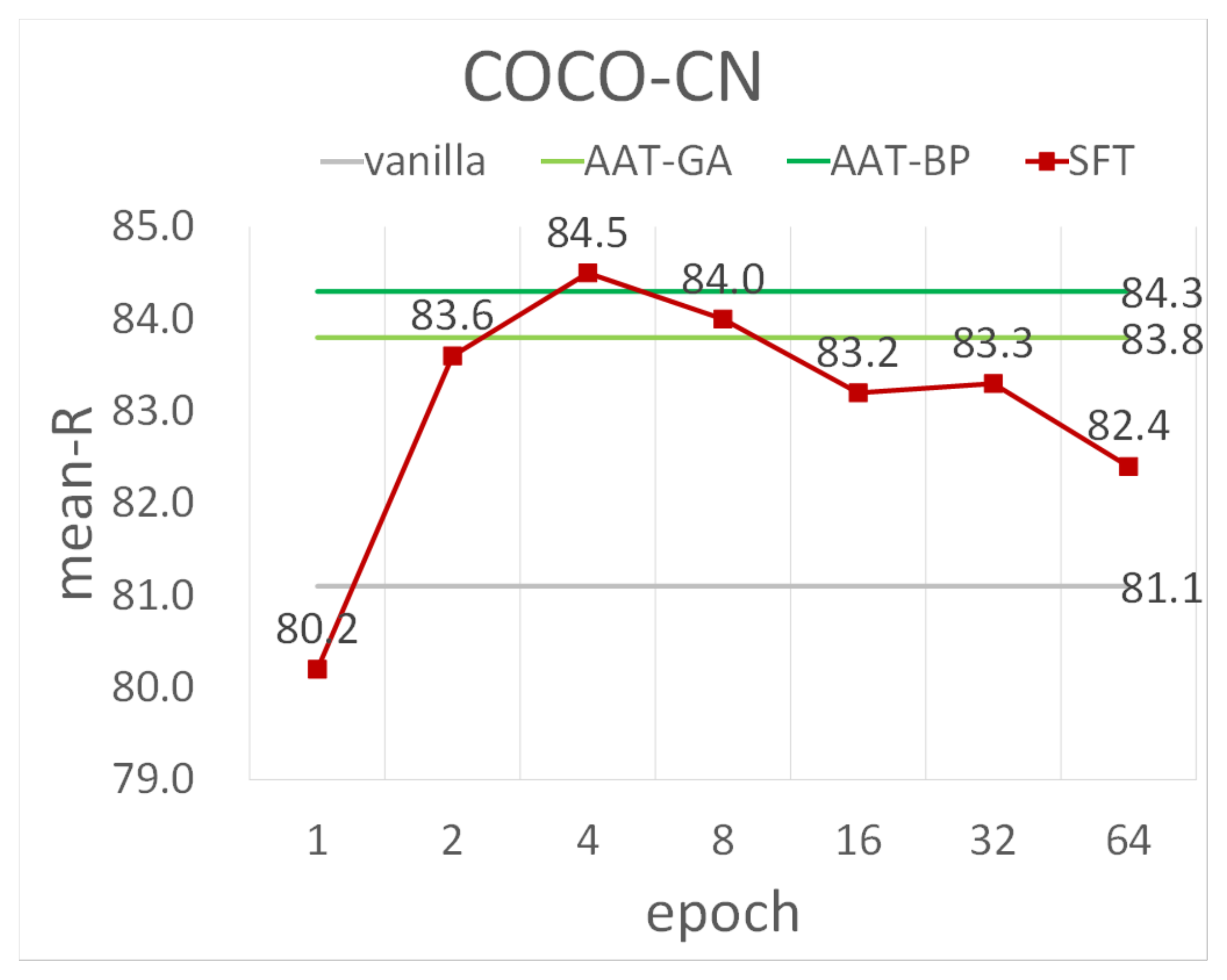}
    \caption{Mean-R on COCO-CN, based on ViT-B.}
    \label{fig:sft-1}
  \end{subfigure}\hfill
  \begin{subfigure}[t]{0.47\textwidth}
    \centering
    \includegraphics[width=\linewidth]{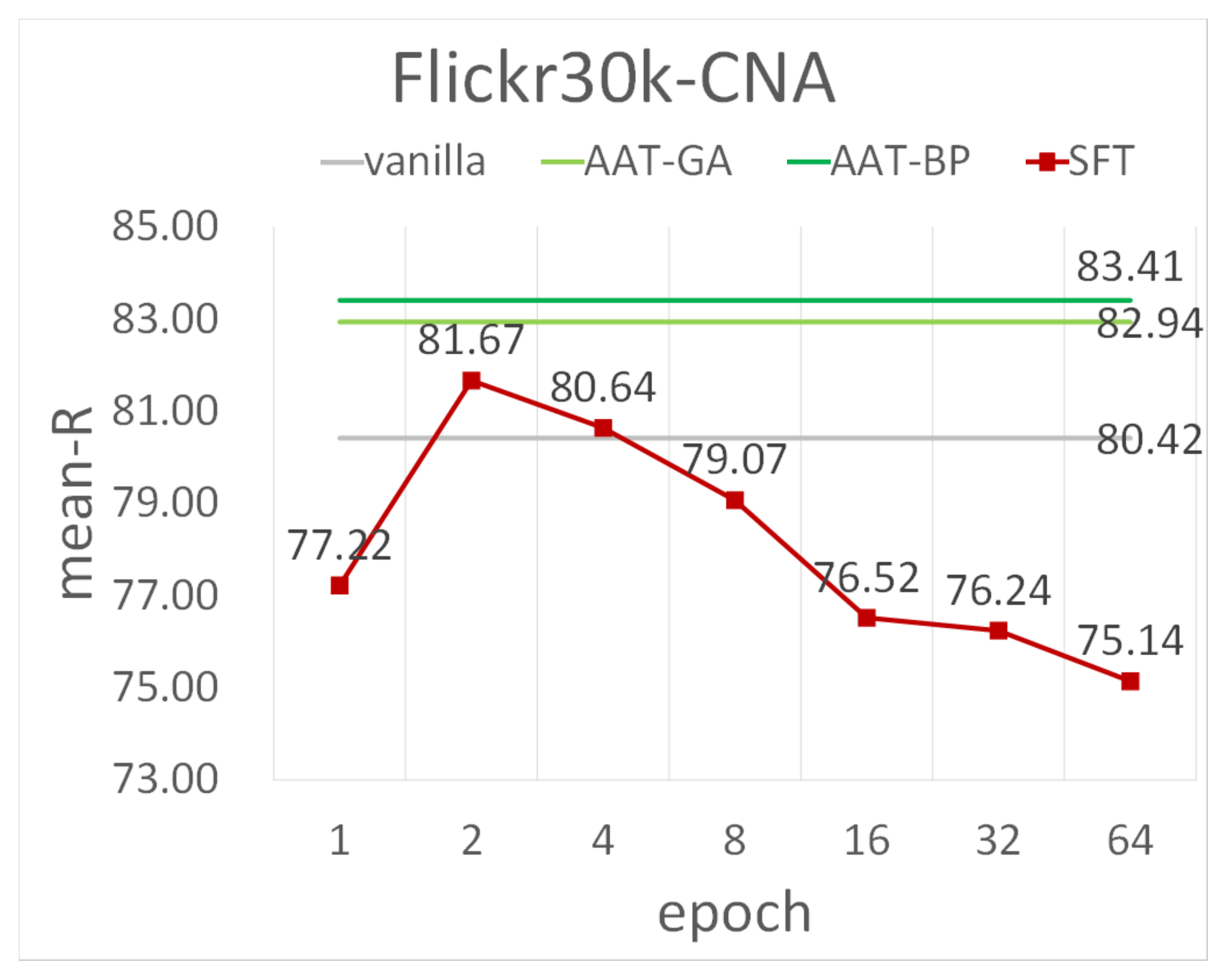}
    \caption{Mean-R on Flickr30k-CNA, based on ViT-B.}
    \label{fig:sft-2}
  \end{subfigure}
  \medskip
  \begin{subfigure}[t]{0.47\textwidth}
    \centering
    \includegraphics[width=\linewidth]{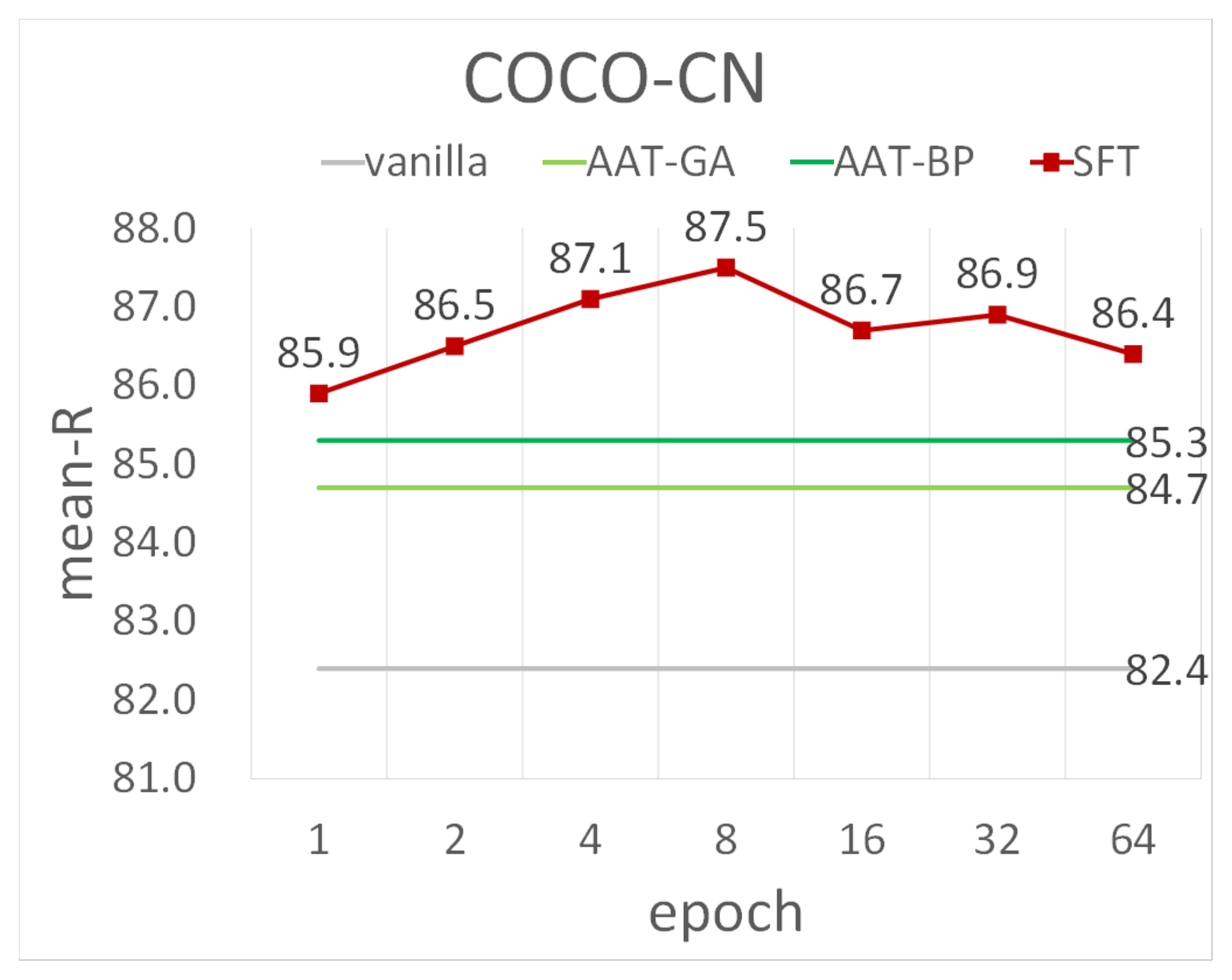}
    \caption{Mean-R on COCO-CN, based on ViT-L.}
    \label{fig:sft-3}
  \end{subfigure}\hfill
  \begin{subfigure}[t]{0.47\textwidth}
    \centering
    \includegraphics[width=\linewidth]{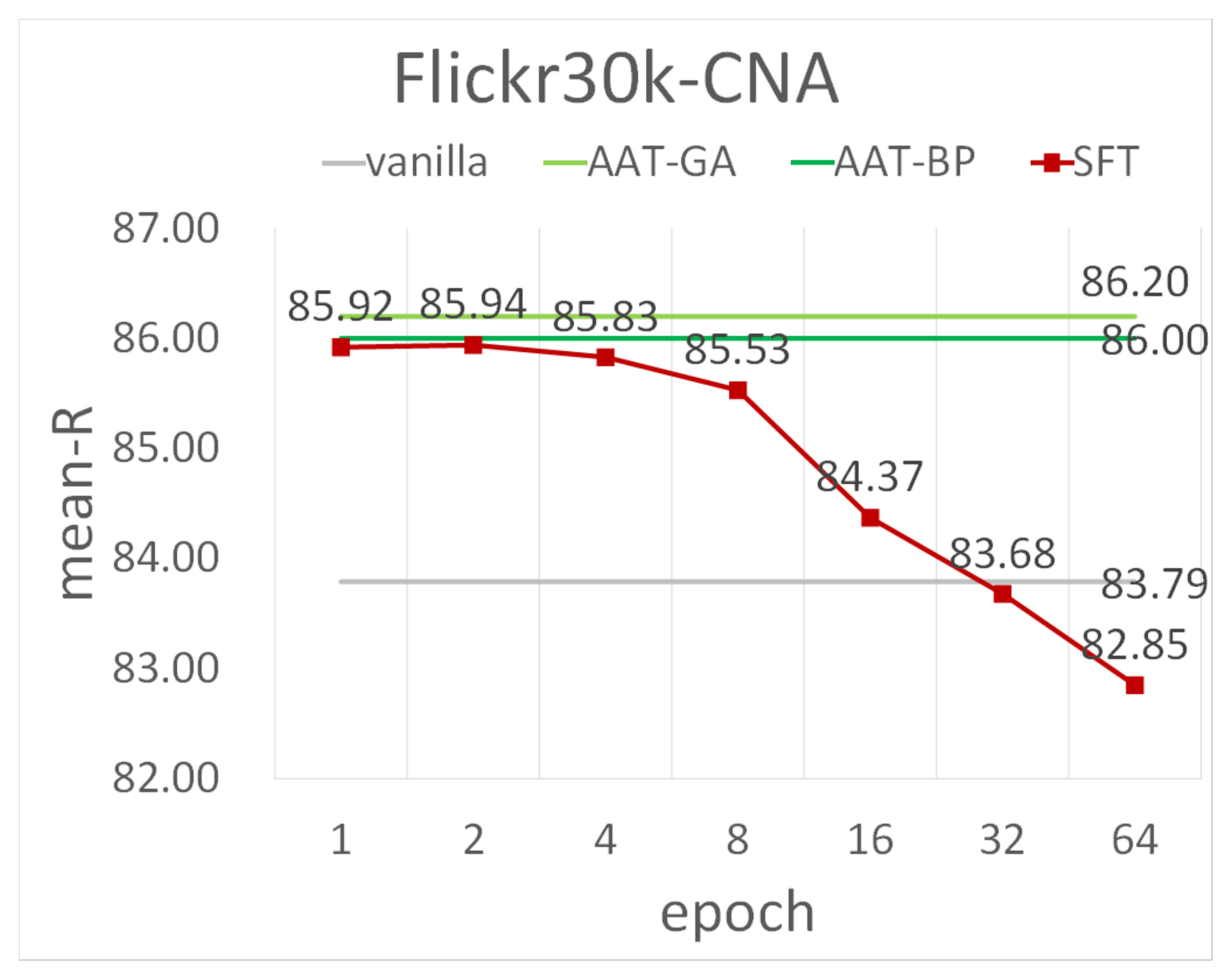}
    \caption{Mean-R on Flickr30k-CNA, based on ViT-L.}
    \label{fig:sft-4}
  \end{subfigure}
  \medskip
  \begin{subfigure}[t]{0.47\textwidth}
    \centering
    \includegraphics[width=\linewidth]{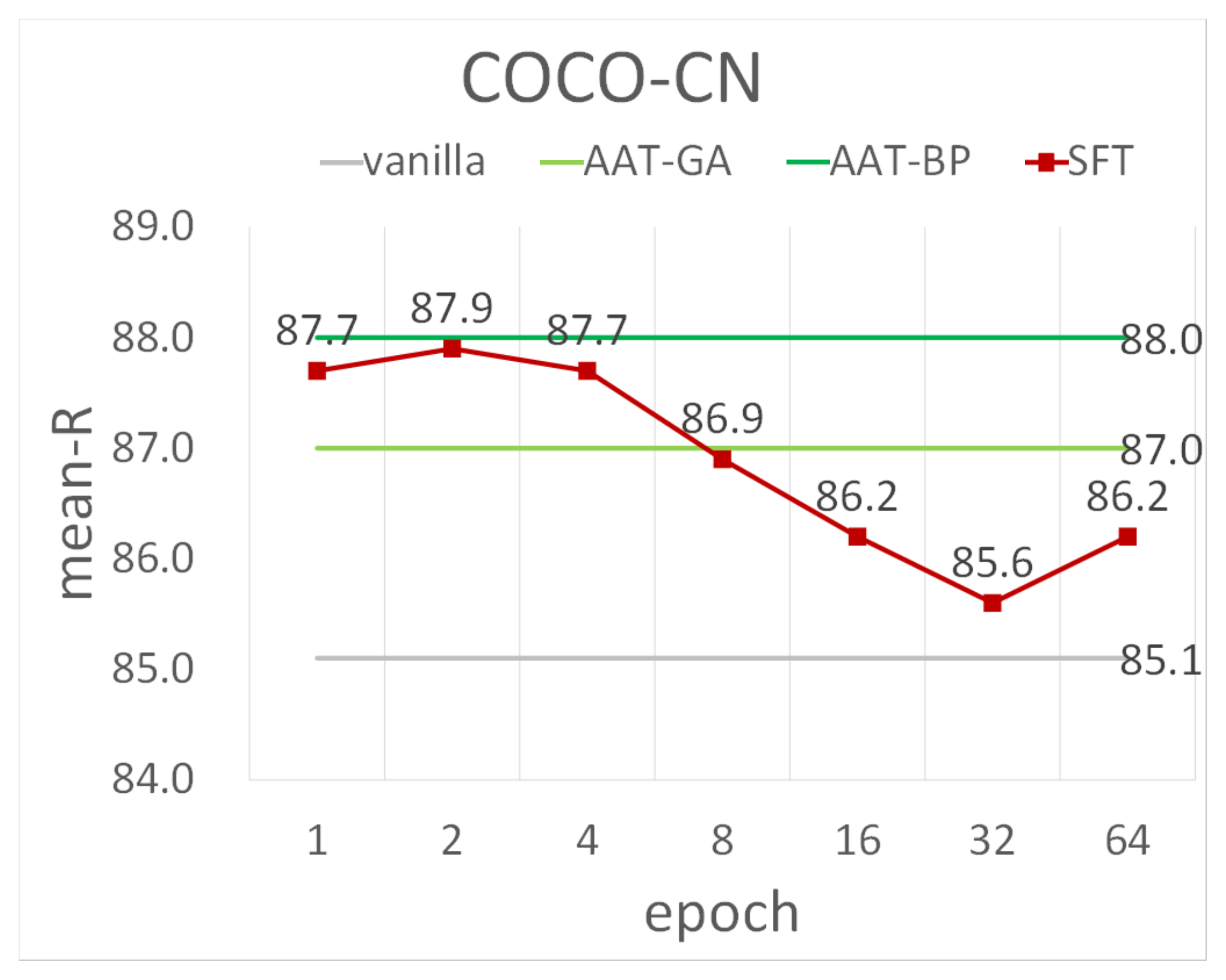}
    \caption{Mean-R on COCO-CN, based on ViT-H.}
    \label{fig:sft-5}
  \end{subfigure}\hfill
  \begin{subfigure}[t]{0.47\textwidth}
    \centering
    \includegraphics[width=\linewidth]{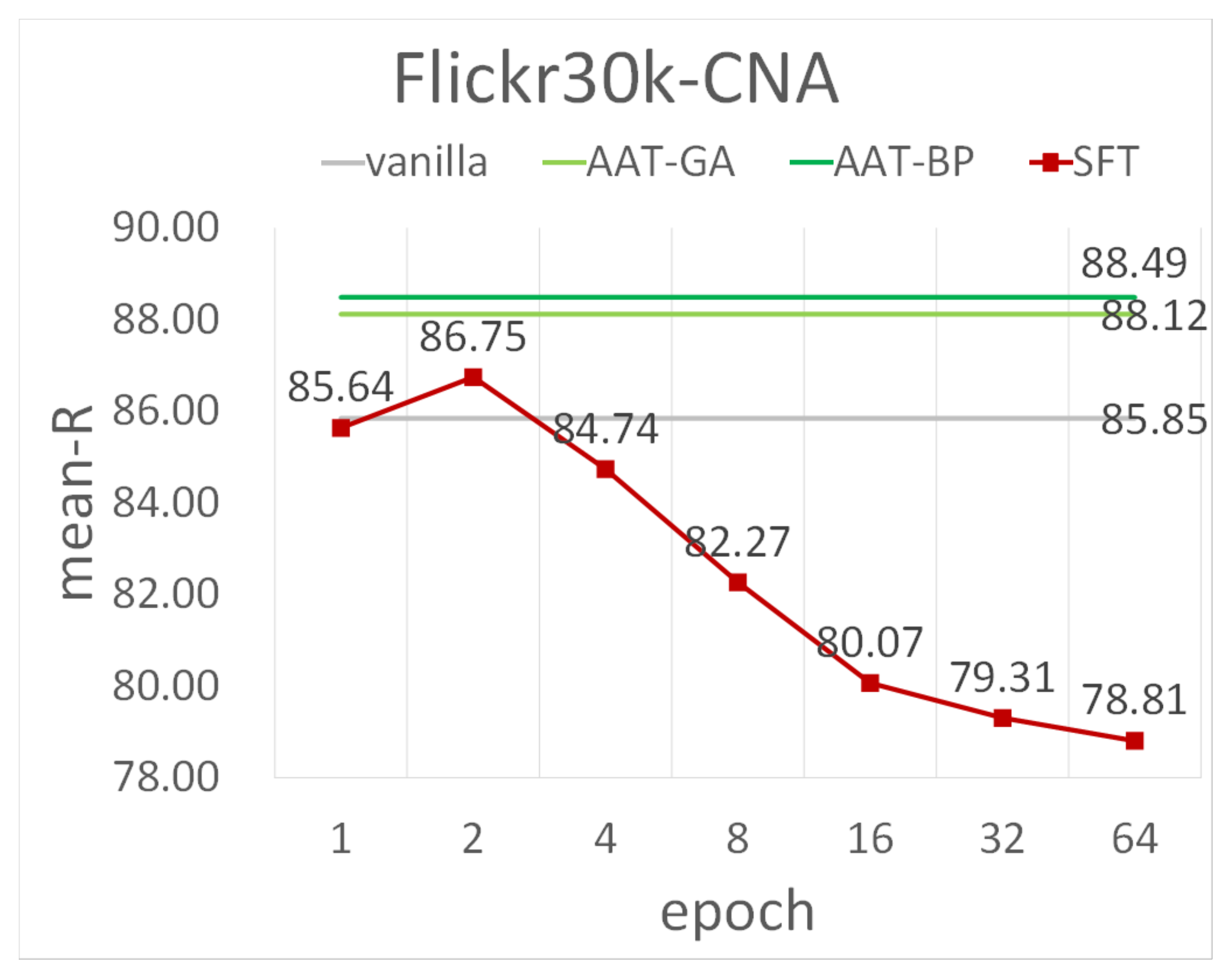}
    \caption{Mean-R on Flickr30k-CNA, based on ViT-H.}
    \label{fig:sft-6}
  \end{subfigure}
  \caption{Comparison for text-to-image retrieval among AAT-improved models, the SFT models, and the vanilla counterparts. For SFT models, mean-R across increasing training epochs are reported. Evaluation is conducted on the \emph{test} sets of COCO-CN and Flickr30k-CNA for each model variant.}
  \label{app:fig:sft}
\end{figure*}

To further validate the effectiveness and practicality of AAT, we compare it against conventional supervised finetuning (SFT). We conduct this comparison using Chinese-CLIP, where AAT demonstrates more pronounced improvements than on the English counterpart, making performance differences easier to observe. Since AAT are optimized using 1k image-text pairs from COCO-CN, we train the vanilla Chinese-CLIP on the same dataset using the contrastive supervision objective employed in CLIP, following the recommended hyper-parameters from the OpenCLIP codebase~\cite{ilharco}. Specifically, we use a base learning rate of 5e-5 with a cosine decay schedule, the AdamW optimizer, a batch size of 256, and a weight decay of 1e-3. In experiments, we vary the number of training epochs from 1 to 64 to explore optimal performance across different model sizes. We evaluate the finetuned models on both the in-domain COCO-CN \emph{test} set and the cross-domain Flickr30k-CNA \emph{test} set for text-to-image retrieval. The results are presented in Figure~\ref{app:fig:sft}.

SFT exhibits varying mean-R on COCO-CN across model sizes. It slightly outperforms AAT on ViT-B ($+$0.2\%), achieves a larger gain on ViT-L ($+$2.2\%), and performs comparably on ViT-H ($-$0.1\%). However, its results are highly sensitive to the number of training epochs, likely due to overfitting or convergence to local minima under limited data. This instability is consistent across repeated trials, highlighting the challenges of SFT in data-scarce settings.

Although SFT beats AAT in some cases, AAT exhibits markedly stronger cross-domain generalization. As shown in the right column of Figure~\ref{app:fig:sft}, AAT consistently outperforms SFT by at least 1.27\% on ViT-B, 0.06\% on ViT-L, and 1.37\% on ViT-H. Notably, the SFT models that perform best on COCO-CN exhibit substantial performance degradation on Flickr30k-CNA, in some cases even falling below the baseline. This indicates a vulnerability of SFT when trained on limited data.

\begin{table}[t]
\centering
\small
\begin{tabular}{p{45pt}<{\centering}p{28pt}<{\centering}p{28pt}<{\centering}p{28pt}<{\centering}p{40pt}<{\centering}}
\hline
 & \multicolumn{3}{c}{single-object} & \multirow{2}{*}{\begin{tabular}[c]{@{}c@{}}multi-object\end{tabular}} \\
\cline{2-4}
 & \multicolumn{1}{c}{GQA} & \multicolumn{1}{c}{CLEVR} & \multicolumn{1}{c}{PACO} & \\
\hline
\multicolumn{1}{c}{vanilla} & 35.36 & \textbf{7.57} & 26.38 & 18.10 \\
\multicolumn{1}{c}{AAT-GA} & \textbf{37.63} & 7.54 & \textbf{28.33} & 20.48 \\
\multicolumn{1}{c}{AAT-BP} & 36.41 & 7.27 & 27.69 & \textbf{21.90} \\
\hline
\end{tabular}
\caption{Comparison of CLIP and AAT-improved CLIP on \emph{Cola}. Results are based on ViT-B, with mAP used as evaluation metric for single-object queries and mean accuracy for multi-object queries.}\label{tab:cola}
\end{table}

\section{Visualization}\label{app:fail}
\subsection{AAT Optimization Data}
For the convenience of readers, we provide a visualization of the validation set $\mathcal{D}$ used for AAT optimization in Figure~\ref{app:fig:coco}.

\subsection{Failure Cases}
\begin{figure*}[!ht]
\centering
\includegraphics[width=6.8in]{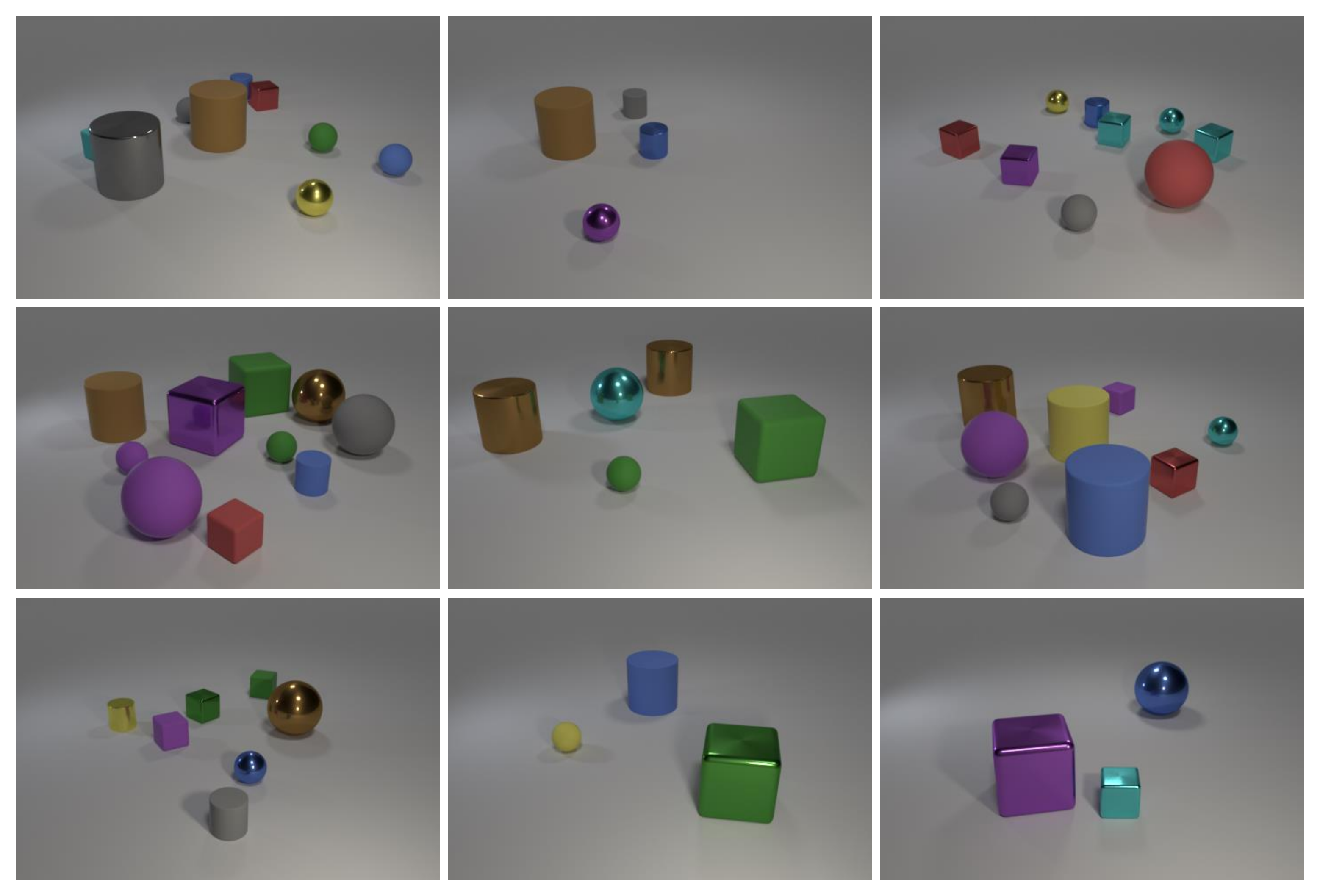}
\caption{Visualization of samples from \emph{Cola}-CLEVR, a subset of \emph{Cola} used as a retrieval benchmark in this paper. Each image is paired with a set of descriptive attributes, such as ``large,'' ``purple,'' ``rubber,'' ``cube,'' ``metal,'' etc.}
\label{app:fig:clevr}
\end{figure*}

\begin{figure*}[htbp]
  \centering
  \begin{subfigure}[t]{0.97\textwidth}
    \centering
    \includegraphics[width=\linewidth]{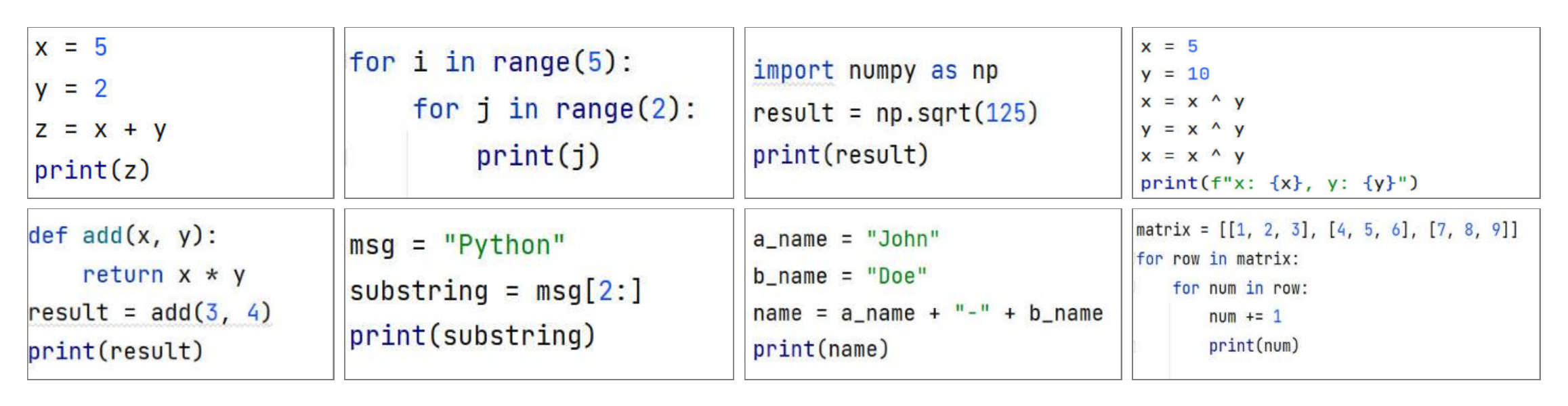}
    \caption{Illustrative examples from the subset of code reasoning within ReCoS-v1. Each image is paired with a text description explaining the reasoning process behind the code presented in the image.}
    \label{fig:recos-cr}
  \end{subfigure}
  \medskip
  \begin{subfigure}[t]{0.955\textwidth}
    \centering
    \includegraphics[width=\linewidth]{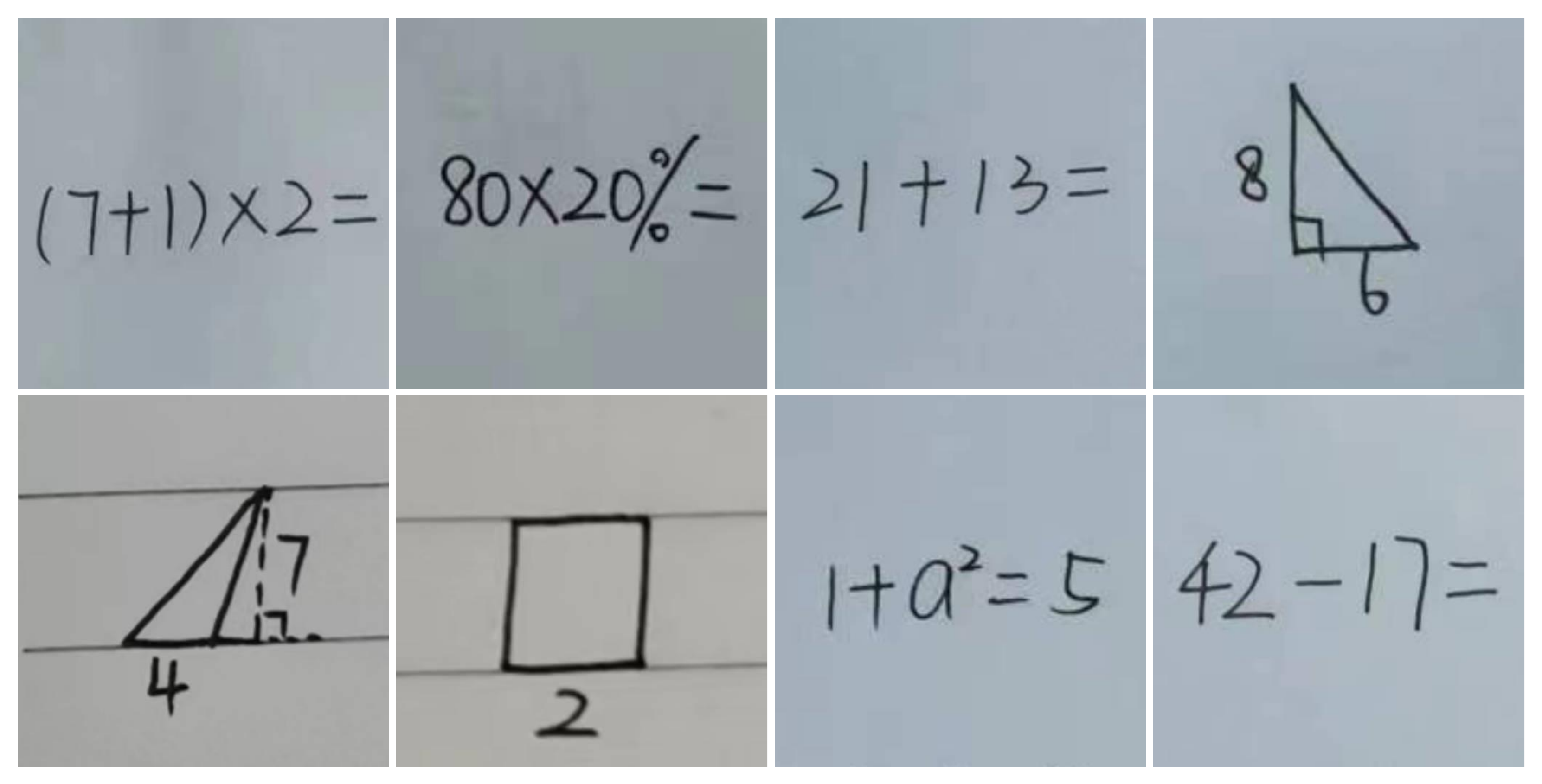}
    \caption{Illustrative examples from the subset of numerical calculation within ReCoS-v1. Each image is paired with a text description that either provides answers to numerical calculations or describes the geometric shapes depicted in the image.}
    \label{fig:recos-nc}
  \end{subfigure}
  \medskip
  \begin{subfigure}[t]{0.955\textwidth}
    \centering
    \includegraphics[width=\linewidth]{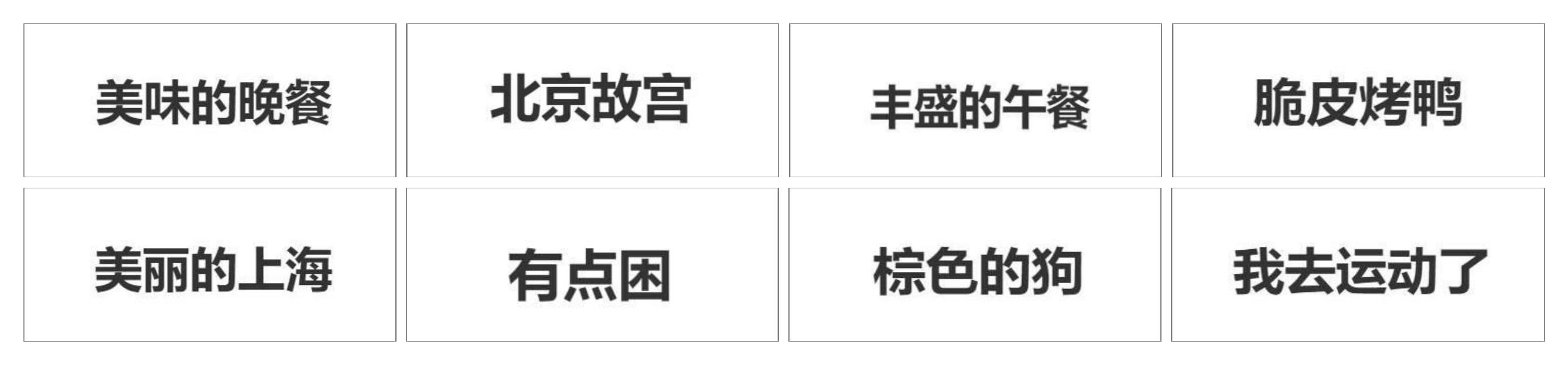}
    \caption{Illustrative examples from the subset of text translation within ReCoS-v1. Each image is paired with a text description that provides the English translation of the Chinese text depicted in the image. For instance, ``The English translation corresponding to the text in the text-only picture is `Delicious dinner','' corresponds to the first image in the top row.}
    \label{fig:recos-tt}
  \end{subfigure}
  \caption{Visualization of non-natural images from three ReCoS-v1 subsets: code reasoning (CR), numerical calculation (NC), and text translation (TT).}
  \label{app:fig:recos}
\end{figure*}

As discussed in Appendix~\ref{app:com} and Section~\ref{sec:res-clip} of the paper, AAT fails on \emph{Cola}-CLEVR and three non-natural subsets from ReCoS. Representative examples from these tasks are shown in Figure~\ref{app:fig:clevr} and Figure~\ref{app:fig:recos}, respectively. As illustrated, there exists a significant domain gap between the test images and those in $\mathcal{D}$. Moreover, certain tasks require models to perform high-level comprehension and reasoning, which exceeds the representational capacity of CLIP.

\section{Limitations}\label{app:limit}
We summarize four limitations of this work, which we leave as directions for future exploration:
\begin{itemize}
\item \textbf{Sample Selection for Optimization:}
It remains unclear how to optimally select the samples used for AAT optimization---specifically, which image-text pairs are most effective for refining the output representations while ensuring broad generalization.
\item \textbf{Dependence on Optimization Data:}
Although AAT is designed for data-scarce scenarios, it still requires a small validation set. Head selection is inherently data-dependent, even when the dataset is indeed minimal.
\item \textbf{Limited Validation on More Challenging Tasks:}
AAT has not yet been evaluated on downstream tasks beyond cross-modal alignment, such as visual question answering (VQA), object detection, or semantic segmentation. Its applicability to these more complex settings remains an open question.
\item \textbf{Lack of Per-Instance Adaptation:}
AAT performs ablation globally based on the entire validation set, without per-instance customization. In practice, certain attention heads may be detrimental in some contexts but beneficial in others. This suggests a promising direction: per-image adaptive head ablation, where head selection is dynamically tailored to the input characteristics.
\end{itemize}

\clearpage

\clearpage
\begin{figure*}[htbp]
\centering
\includegraphics[width=6.75in]{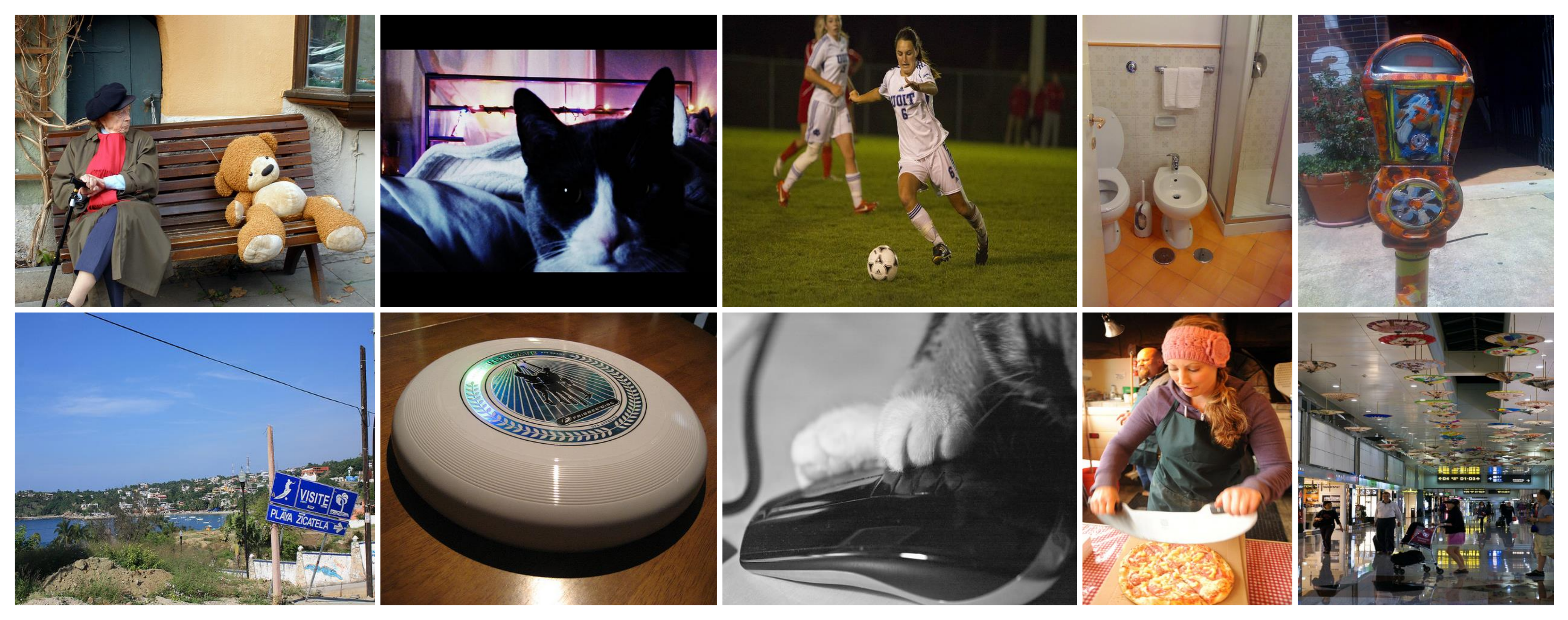}
\caption{Visualization of samples from the validation set $\mathcal{D}$ used for AAT optimization, sourced from MS COCO. Each image is paired with a single text description, such as, ``An old woman sits next to a large stuffed teddy bear on a bench,'' which corresponds to the first image in the top row.}
\label{app:fig:coco}
\end{figure*}

\clearpage

\clearpage
\begin{table*}[htbp]
\centering
\small
\begin{tabular}{p{10pt}<{\centering}|p{10pt}<{\centering}|p{16pt}<{\centering}p{16pt}<{\centering}p{16pt}<{\centering}p{16pt}<{\centering}|p{10pt}<{\centering}|p{10pt}<{\centering}|p{16pt}<{\centering}p{16pt}<{\centering}p{16pt}<{\centering}p{16pt}<{\centering}|p{10pt}<{\centering}|p{10pt}<{\centering}|p{16pt}<{\centering}p{16pt}<{\centering}p{16pt}<{\centering}p{16pt}<{\centering}}
\hline
$l$ & $h$ & R@1 & R@5 & R@10 & mR & $l$ & $h$ & R@1 & R@5 & R@10 & mR & $l$ & $h$ & R@1 & R@5 & R@10 & mR \\
\hline
\multirow{12}{*}{0} & 0 & 60.8 & 86.0 & 93.5 & 80.1 & \multirow{12}{*}{1} & 0 & 60.3 & 86.3 & 93.7 & 80.1 & \multirow{12}{*}{2} & 0 & 60.8 & 86.3 & 93.3 & 80.1 \\
 & 1 & 59.9 & 86.3 & 93.4 & 79.9 & & 1 & 60.7 & 86.2 & 93.2 & 80.0 & & 1 & 60.4 & 86.1 & 93.2 & 79.9 \\
 & 2 & 60.4 & 86.2 & 93.2 & 79.9 & & 2 & 60.9 & 86.4 & 93.0 & 80.1 & & 2 & 60.1 & 86.3 & 93.2 & 79.9  \\
 & 3 & 60.2 & 86.5 & 93.3 & 80.0 & & 3 & 60.6 & 85.7 & 93.3 & 79.9 & & 3 & 60.1 & 85.9 & 93.3 & 79.8 \\
 & 4 & 60.8 & 86.3 & 93.5 & 80.2 & & 4 & 60.2 & 85.8 & 93.4 & 79.8 & & 4 & 60.3 & 86.0 & 93.3 & 79.9 \\
 & 5 & 60.2 & 86.3 & 93.3 & 79.9 & & 5 & 60.3 & 86.1 & 93.2 & 79.9 & & 5 & 60.7 & 85.9 & 93.2 & 79.9 \\
 & 6 & 60.5 & 86.1 & 93.4 & 80.0 & & 6 & 60.4 & 86.2 & 93.2 & 79.9 & & 6 & 60.3 & 85.8 & 93.4 & 79.8 \\
 & 7 & 60.0 & 86.2 & 93.6 & 79.9 & & 7 & 60.2 & 86.4 & 93.2 & 79.9 & & 7 & 58.2 & 85.1 & 93.0 & 78.8 \\
 & 8 & 60.4 & 86.4 & 93.2 & 80.0 & & 8 & 60.4 & 86.3 & 93.3 & 80.0 & & 8 & 60.4 & 86.6 & 93.1 & 80.0 \\
 & 9 & 60.3 & 86.2 & 93.2 & 79.9 & & 9 & 60.6 & 86.3 & 93.2 & 80.0 & & 9 & 60.5 & 86.1 & 93.2 & 79.9 \\
 & 10 & 60.4 & 86.4 & 93.2 & 80.0 & & 10 & 60.2 & 86.1 & 93.4 & 79.9 & & 10 & 60.3 & 86.3 & 93.4 & 80.0 \\
 & 11 & 60.2 & 85.9 & 93.6 & 79.9 & & 11 & 60.4 & 86.0 & 93.2 & 79.9 & & 11 & 60.7 & 86.5 & 93.0 & 80.1 \\
\hline
\multirow{12}{*}{3} & 0 & 60.9 & 85.7 & 93.6 & 80.1 & \multirow{12}{*}{4} & 0 & 60.6 & 86.4 & 93.4 & 80.1 & \multirow{12}{*}{5} & 0 & 60.8 & 85.8 & 93.4 & 80.0 \\
 & 1 & 60.1 & 86.1 & 93.1 & 79.8 & & 1 & 59.3 & 85.9 & 93.2 & 79.5 & & 1 & 60.4 & 86.2 & 93.4 & 80.0 \\
 & 2 & 59.1 & 85.8 & 92.6 & 79.2 & & 2 & 60.4 & 85.8 & 93.2 & 79.8 & & 2 & 60.6 & 86.1 & 93.4 & 80.0 \\
 & 3 & 60.0 & 86.2 & 93.3 & 79.8 & & 3 & 60.0 & 86.4 & 93.1 & 79.8 & & 3 & 58.9 & 84.4 & 92.4 & 78.6 \\
 & 4 & 59.6 & 85.9 & 93.4 & 79.6 & & 4 & 60.2 & 85.8 & 93.4 & 79.8 & & 4 & 60.6 & 86.5 & 93.3 & 80.1 \\
 & 5 & 60.3 & 86.6 & 93.1 & 80.0 & & 5 & 59.9 & 86.3 & 93.3 & 79.8 & & 5 & 60.2 & 86.0 & 93.3 & 79.8 \\
 & 6 & 60.5 & 86.3 & 93.2 & 80.0 & & 6 & 59.7 & 86.4 & 93.3 & 79.8 & & 6 & 59.9 & 86.2 & 93.3 & 79.8 \\
 & 7 & 60.5 & 86.3 & 93.4 & 80.1 & & 7 & 61.1 & 86.0 & 93.3 & 80.1 & & 7 & 60.6 & 86.4 & 93.6 & 80.2 \\
 & 8 & 60.7 & 86.5 & 93.3 & 80.2 & & 8 & 60.4 & 86.4 & 93.4 & 80.1 & & 8 & 60.1 & 85.7 & 93.4 & 79.7 \\
 & 9 & 60.1 & 86.5 & 93.3 & 80.0 & & 9 & 59.1 & 85.9 & 93.2 & 79.4 & & 9 & 60.3 & 86.5 & 93.1 & 80.0 \\
 & 10 & 59.7 & 86.8 & 93.2 & 79.9 & & 10 & 60.3 & 86.4 & 93.5 & 80.1 & & 10 & 60.0 & 85.4 & 93.1 & 79.5 \\
 & 11 & 57.4 & 83.3 & 91.0 & 77.2 & & 11 & 60.2 & 86.1 & 92.9 & 79.7 & & 11 & 61.0 & 86.0 & 93.5 & 80.2 \\
\hline
\multirow{12}{*}{6} & 0 & 60.4 & 86.3 & 93.1 & 79.9 & \multirow{12}{*}{7} & 0 & 60.0 & 85.8 & 93.4 & 79.7 & \multirow{12}{*}{8} & 0 & 60.4 & 86.5 & 93.3 & 80.1 \\
 & 1 & 59.4 & 86.2 & 93.4 & 79.7 & & 1 & 60.1 & 86.2 & 93.3 & 79.9 & & 1 & 59.8 & 86.0 & 92.9 & 79.6 \\
 & 2 & 59.9 & 85.3 & 92.9 & 79.4 & & 2 & 61.0 & 86.7 & 93.3 & 80.3 & & 2 & 55.0 & 83.7 & 91.2 & 76.6 \\
 & 3 & 60.4 & 86.2 & 93.1 & 79.9 & & 3 & 59.3 & 86.0 & 92.8 & 79.4 & & 3 & 60.5 & 86.1 & 93.4 & 80.0 \\
 & 4 & 60.6 & 86.3 & 93.4 & 80.1 & & 4 & 60.2 & 86.2 & 92.9 & 79.8 & & 4 & 79.9 & 85.9 & 93.3 & 79.9 \\
 & 5 & 59.9 & 86.4 & 93.2 & 79.8 & & 5 & 60.5 & 86.2 & 93.3 & 80.0 & & 5 & 60.3 & 86.2 & 93.4 & 80.0 \\
 & 6 & 59.5 & 86.3 & 93.4 & 79.7 & & 6 & 59.9 & 86.2 & 92.9 & 79.7 & & 6 & 60.7 & 85.7 & 92.8 & 79.7 \\
 & 7 & 60.3 & 86.2 & 93.1 & 79.9 & & 7 & 60.3 & 86.5 & 93.1 & 80.0 & & 7 & 60.4 & 86.4 & 93.0 & 79.9 \\
 & 8 & 60.3 & 86.1 & 93.5 & 80.0 & & 8 & 60.6 & 86.2 & 93.2 & 80.0 & & 8 & 60.0 & 86.7 & 92.8 & 79.8 \\
 & 9 & 60.7 & 86.2 & 93.2 & 80.0 & & 9 & 60.1 & 85.9 & 93.1 & 79.7 & & 9 & 59.9 & 86.4 & 93.3 & 79.9 \\
 & 10 & 60.3 & 86.2 & 93.2 & 79.9 & & 10 & 59.8 & 86.6 & 93.3 & 79.9 & & 10 & 61.0 & 86.2 & 93.6 & 80.3 \\
 & 11 & 60.8 & 86.7 & 93.8 & 80.4 & & 11 & 59.9 & 86.2 & 93.2 & 79.8 & & 11 & 60.1 & 86.1 & 93.1 & 79.8 \\
\hline
\multirow{12}{*}{9} & 0 & 59.5 & 86.2 & 93.0 & 79.6 & \multirow{12}{*}{10} & 0 & 60.2 & 86.7 & 93.4 & 80.1 & \multirow{12}{*}{11} & 0 & 61.4 & 87.4 & 93.4 & 80.7 \\
 & 1 & 60.8 & 86.4 & 93.3 & 80.2 & & 1 & 60.6 & 86.0 & 93.0 & 79.9 & & 1 & 54.7 & 83.0 & 91.0 & 76.2 \\
 & 2 & 60.4 & 86.8 & 93.4 & 80.2 & & 2 & 60.7 & 85.6 & 93.3 & 79.9 & & 2 & 59.3 & 84.9 & 91.9 & 78.7 \\
 & 3 & 60.1 & 86.4 & 93.2 & 79.9 & & 3 & 59.6 & 86.6 & 93.0 & 79.7 & & 3 & 59.4 & 83.5 & 90.9 & 77.9 \\
 & 4 & 59.8 & 86.2 & 92.7 & 79.6 & & 4 & 61.0 & 86.9 & 93.4 & 80.4 & & 4 & 60.2 & 86.2 & 93.2 & 79.9 \\
 & 5 & 59.7 & 86.7 & 92.8 & 79.7 & & 5 & 60.6 & 85.6 & 93.9 & 80.0 & & 5 & 59.9 & 86.2 & 93.4 & 79.8 \\
 & 6 & 60.8 & 87.2 & 93.4 & 80.5 & & 6 & 60.6 & 86.8 & 93.2 & 80.2 & & 6 & 61.5 & 87.1 & 94.6 & 81.1 \\
 & 7 & 59.4 & 85.3 & 93.4 & 79.4 & & 7 & 62.1 & 86.8 & 93.5 & 80.8 & & 7 & 58.3 & 84.4 & 92.4 & 78.4 \\
 & 8 & 61.2 & 86.5 & 93.5 & 80.4 & & 8 & 60.9 & 86.7 & 93.4 & 80.3 & & 8 & 58.1 & 84.6 & 92.3 & 78.3 \\
 & 9 & 62.0 & 86.8 & 93.4 & 80.7 & & 9 & 61.4 & 87.5 & 93.4 & 80.8 & & 9 & 56.4 & 84.6 & 91.0 & 77.3 \\
 & 10 & 60.4 & 86.7 & 93.4 & 80.2 & & 10 & 61.1 & 86.2 & 93.4 & 80.2 & & 10 & 60.6 & 86.7 & 93.6 & 80.3 \\
 & 11 & 60.0 & 86.8 & 93.3 & 80.0 & & 11 & 61.6 & 86.8 & 93.1 & 80.5 & & 11 & 59.2 & 85.0 & 91.9 & 78.7 \\
\hline
\multicolumn{2}{c|}{vanilla} & 60.4 & 86.2 & 93.2 & 79.9 \\
\hline
\end{tabular}
\value{7.5pt}
\caption{Detailed text-to-image retrieval performance on the COCO-CN \emph{val} set for each individually ablated head, evaluated using the ViT-B-based Chinese-CLIP model. Here, $l$ denotes the $l$-th transformer layer of the image encoder, and $h$ represents the $h$-th attention head in the multi-head attention (MHA) module. Evaluation metrics include recall rates at k (R@k) and mean recall rate (mR). Results from the vanilla baseline are reported in the last row for reference.}\label{app:tab:head}
\end{table*}

\end{document}